\documentclass[sageh,times]{sageh}
\usepackage{longtable}
\usepackage{moreverb,url}
\usepackage{multicol}
\usepackage{multirow}
\usepackage{makecell}
\usepackage{bbm}
\usepackage[table]{xcolor}
\usepackage[colorlinks,bookmarksopen,bookmarksnumbered,citecolor=red,urlcolor=red]{hyperref}
\usepackage{amsmath}
\usepackage{natbib}
\usepackage{pifont}
\usepackage{threeparttable}
\usepackage{float}
\setcitestyle{numbers,square}
\newcommand\BibTeX{{\rmfamily B\kern-.05em \textsc{i\kern-.025em b}\kern-.08em
T\kern-.1667em\lower.7ex\hbox{E}\kern-.125emX}}

\def\volumeyear{2016}

\begin{document}

\runninghead{Lin et al.}

\title{Visuomotor Grasping with World Models for Surgical Robots}

\author{Hongbin Lin, Bin Li, and Kwok Wai Samuel Au}

\affiliation{\affilnum{1}Chinese University of Hong Kong}

\corrauth{Hongbin Lin, Kwok Wai Samuel Au}

\email{hongbinlin@link.cuhk.edu.hk}

\begin{abstract}

Grasping is a fundamental task in robot-assisted surgery (RAS), and automating it can reduce surgeon workload while enhancing efficiency, safety, and consistency beyond teleoperated systems. 
Most prior approaches rely on explicit object pose tracking or handcrafted visual features, limiting their generalization to novel objects, robustness to visual disturbances, and the ability to handle deformable objects.
Visuomotor learning offers a promising alternative, but deploying it in RAS presents unique challenges, such as low signal-to-noise ratio in visual observations, demands for high safety and millimeter-level precision, as well as the complex surgical environment.
This paper addresses three key challenges: 
(i) sim-to-real transfer of visuomotor policies to ex vivo surgical scenes, 
(ii) visuomotor learning using only a single stereo camera pair---the standard RAS setup, and 
(iii) object-agnostic grasping with a single policy that generalizes to diverse, unseen surgical objects without retraining or task-specific models.
We introduce Grasp Anything for Surgery V2 (GASv2), a visuomotor learning framework for surgical grasping. 
GASv2 leverages a world-model-based architecture and a surgical perception pipeline for visual observations, combined with a hybrid control system for safe execution.
We train the policy in simulation using domain randomization for sim-to-real transfer and deploy it on a real robot in both phantom-based and ex vivo surgical settings, using only a single pair of endoscopic cameras.
Extensive experiments show our policy achieves a 65\% success rate in both settings, generalizes to unseen objects and grippers, and adapts to diverse disturbances, demonstrating strong performance, generality, and robustness.

\end{abstract}


\maketitle
\section{Introduction}
\label{sec:intro}

\begin{figure}[!tbp]
  \centering
  \includegraphics[width=1.0\hsize]{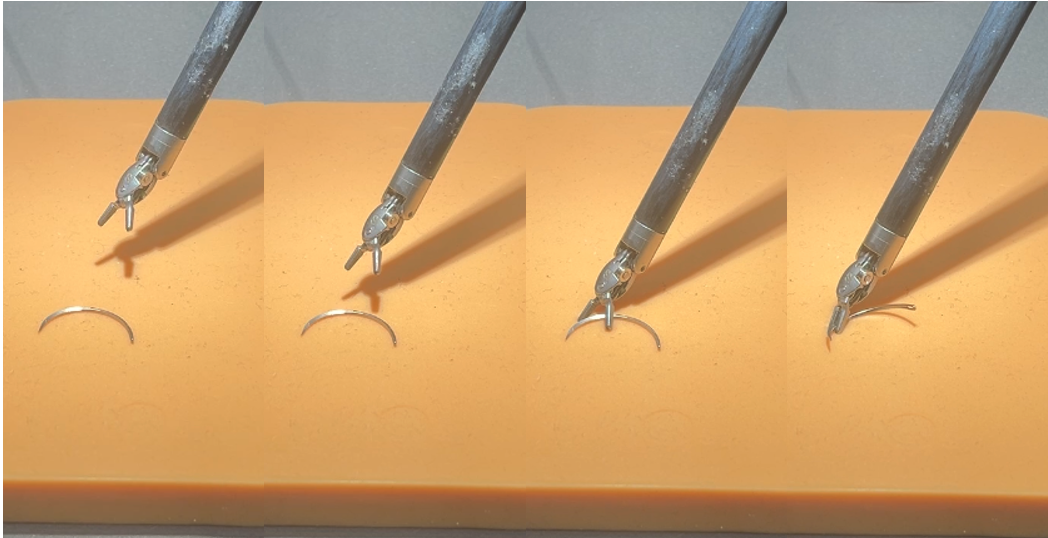}
  \caption{Illustration of surgical grasping task. An agent needs to control the robot gripper to grasp a target object. Typical behaviors of surgical grasping are approaching the target object and closing the gripper.}
  \vspace{-0.45cm}
 \label{fig:grasping_task}
\end{figure}

Grasping is a fundamental and frequent action in surgery, essential for tasks such as handling needles (see Fig.~\ref{fig:grasping_task}), gauze, thread, sponges, and manipulating tissue during procedures like debridement.
In modern robot-assisted surgery (RAS), these tasks are executed via teleoperation, where a surgeon controls the robotic gripper to replicate hand movements.
However, teleoperation demands significant physical and cognitive effort due to the repetitive and tedious nature of grasping, contributing to surgeon fatigue.
Moreover, outcomes vary widely depending on individual surgeon skill and experience \cite{kim2025srt}.
Automating surgical grasping could reduce surgeon workload and improve efficiency, safety, and consistency beyond current teleoperated systems.

Researchers have extensively explored surgical grasping automation using pose-based methods \cite{xu2021surrol} and visual-feature-based methods \cite{zhong2019dual}.
These approaches track low-dimensional spatial states---such as object poses or handcrafted image-space feature coordinates--from visual input and feed them into downstream controllers.
However, they face three key limitations.
First, they often rely on full geometric models or assumed shapes, which limits generalization to unseen objects and introduces assumed geometric bias.
Second, state estimation is sensitive to visual disturbances, including occlusion, fogging, and blood.
Third, low-dimensional states struggle to represent deformable object configurations.
Some approaches use robot proprioception, such as joint measurements, to infer gripper pose without vision.
Yet, most surgical robots (e.g., da Vinci Surgical System) suffer from inaccurate forward kinematics due to potentiometer-based sensing, hysteresis, and mechanical slack, which degrades control accuracy \cite{kim2024surgical}.

Visuomotor learning offers a promising alternative to pose-based and visual-feature-based methods, showing strong performance in general robotic grasping \cite{wu2023daydreamer,kalashnikov2018qt}.
These methods learn end-to-end policies that map visual observations directly to robot actions, implicitly capturing spatial features that are often richer than explicit representations, such as pose or hand-craft visual features.
As a result, visuomotor approaches demonstrate improved generalization to out-of-distribution (OOD) objects \cite{kalashnikov2018qt}, enhanced robustness to visual noise \cite{levine2016end}, and the ability to handle deformable objects \cite{scheikl2022sim}.

Despite the success of visuomotor learning in general robotic grasping, applying it to RAS faces several domain-specific challenges.
First, in typical RAS camera settings, the task-related pixels on the observed image are extremely scarce, with the ratio of task object pixels (i.e., signal) to background pixels (i.e., noise) being approximately 1:99 \cite{lin2024world}.
Such a low signal-to-noise ratio poses significant challenges for training the policy's encoder.
Second, surgical grasping requires millimeter-level precision and safe trajectories to prevent tissue damage, unlike general grasping with coarser centimeter-scale tolerances and lower safety demands.
Third, surgical environments are significantly more complex, with patient-specific anatomy, dynamic scene changes, and frequent visual occlusions (e.g., blood, smoke), as well as unpredictable tissue dynamics, all of which challenge both perception and control \cite{kim2025srt}.

Only a few works have explored visuomotor learning in RAS. 
A transformer-based policy trained via imitation learning has been applied to surgical scenes in both phantom \cite{kim2024surgical} and ex vivo settings \cite{kim2025srt}, but it requires extensive real-robot data collection, incurring high labor and operational costs.
Moreover, it relies on additional wrist-mounted cameras, increasing hardware complexity beyond standard RAS setups (which typically use only endoscopic cameras) and raising concerns about feasibility in in vivo settings due to port constraints.
To circumvent real-robot data collection, Scheikl et al. train a visuomotor policy in simulation using PPO and transfer it to phantom scenes via pixel-level domain randomization \cite{scheikl2022sim}.
However, its applicability to ex vivo settings remains unclear, especially given its reliance on a RealSense depth camera, which may not be suitable for surgical environments.

\begin{figure*}[!tbp]
  \centering
  \includegraphics[width=1.0\hsize]{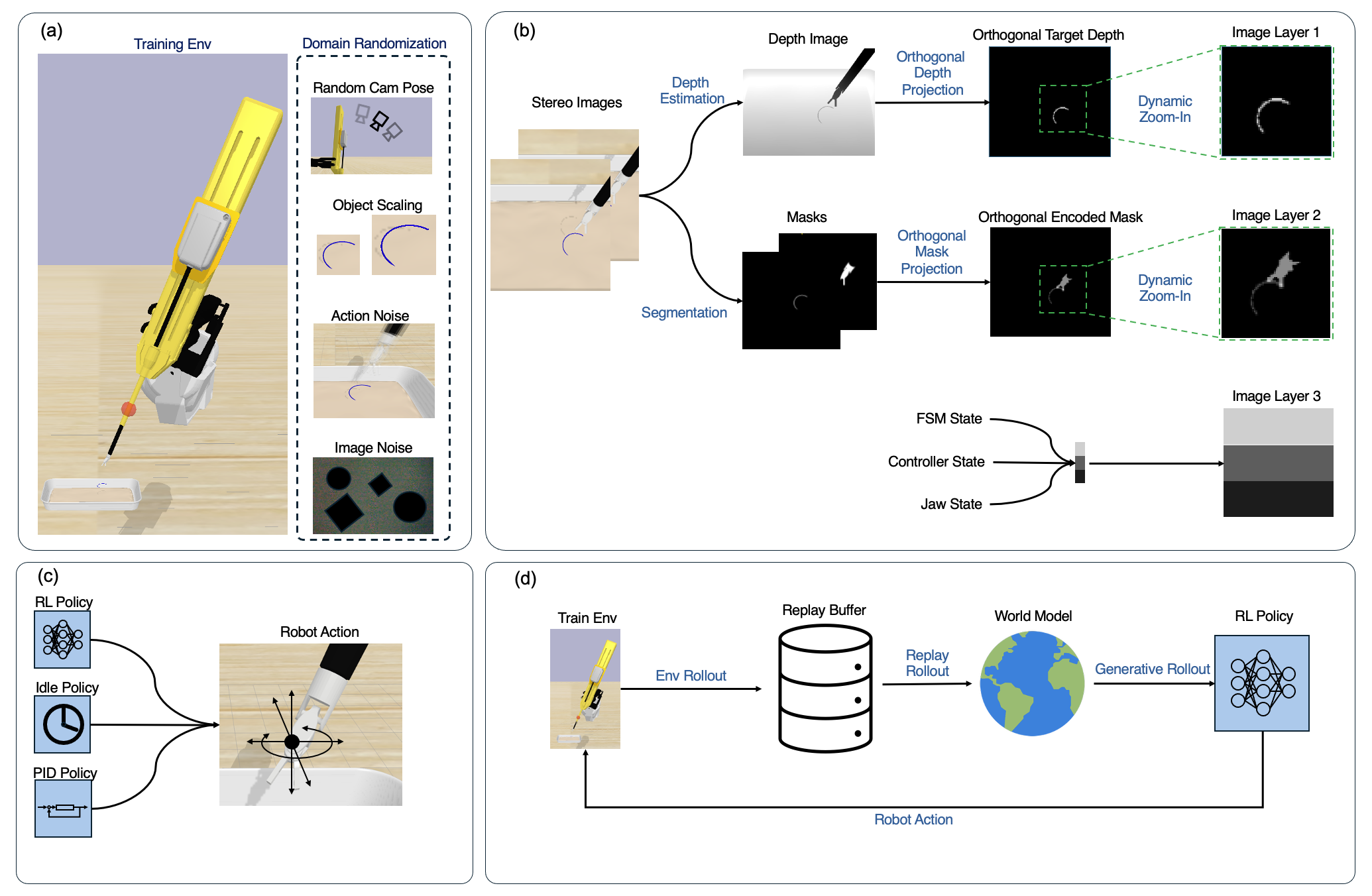}
  \caption{Overview of GASv2. 
 It consists of four components: (a) training scenes with domain randomization, (b) a surgical perception pipeline, (c) a hybrid control architecture, and (d) world-model-based visuomotor learning.}
  \vspace{-0.45cm}
 \label{fig:gasv2_overview}
\end{figure*}

In this paper, we aim to address three unsolved challenges beyond prior work: 
(i) sim-to-real transfer of visuomotor policies to ex vivo surgical scenes, 
(ii) visuomotor learning using only a single stereo camera pair---the standard RAS setup, and 
(iii) object-agnostic grasping with a single policy that generalizes to diverse, unseen surgical objects without retraining or task-specific models.
To this end, we propose a visuomotor learning framework, \textit{\textbf{G}rasp \textbf{A}nything for \textbf{S}urgery V2} (GASv2), for surgical grasping (see Fig.~\ref{fig:gasv2_overview}).
GASv2 adopts a world-model-based architecture to learn a visuomotor policy.
A surgical perception pipeline is leveraged for visual observations of the policy.
A hybrid control system integrates the learned policy with traditional controllers for safe and precise execution.
We train the policy in simulation with domain randomization for sim-to-real transfer, and deploy it on a real robot in both phantom-based and ex vivo surgical scenes---using only a single pair of endoscopic cameras, consistent with standard RAS settings.
In summary, our primary contributions are:
\color{black}
\begin{enumerate}
  \item A data-driven framework for sim-to-real learning of object-agnostic visuomotor policy for surgical grasping.
  We present the first successful deployment of a visuomotor policy transferred from simulation to an ex vivo setting, using only visual input from a standard pair of endoscopic cameras.

\item A novel image representation method for visuomotor policy learning in surgical grasping.  
Our approach encodes depth images, segmentation masks, and system states into a compact $64 \times 64 \times 3$ image.  
It preserves high-resolution detail in task-critical local regions via dynamic zooming, while meeting image size constraints imposed by the limited capacity of our world-model-based policy.

\item Extensive experiments in simulation and on real robots to evaluate the performance, generalization, and robustness of our learned visuomotor policy.  
Our policy achieves strong performance, with a 65\% success rate in both phantom-based and ex vivo scenes.  
It generalizes to unseen objects and grippers, and adapts to diverse disturbances, demonstrating its generality and robustness.

\end{enumerate}


This work is an extended version of previously published conference papers \cite{lin2023end,lin2024world}.
Compared to our previous works, we propose additional methods and significantly expand the experimental evaluation of the overall methodology in the following ways:
\begin{enumerate}
    \setcounter{enumi}{2}
    \item Additional modules in our surgical perception pipeline, including a voxel-based orthogonal projection for equivariant image representation and a depth estimation network for stereo images. 
    \item A hybrid control architecture that integrates traditional control policies with the learned visuomotor policy to improve the sampling efficiency of RL training, reduce corner cases, and ensure low error of hidden state estimation. 

    \item Experiments in novel surgical scenes, including a multi-level phantom setup to evaluate effectiveness under varying object heights, and a porcine stomach for ex vivo evaluation.

\end{enumerate}

\section{Related Works}
\subsection{Autonomous Surgical Grasping}

Previous studies have extensively explored object grasping in surgical tasks, including needle grasping, gauze picking, thread manipulation, debridement, and peg transfer.
Among these, needle grasping is the most widely studied task in surgical robotics.
For the needle-grasping task, robots are required to either pick up a needle or re-grasp a needle held by another gripper.
A traditional way to perform needle grasping is first tracking the needle pose visually and then planning feasible trajectories based on the tracked pose via analytical solutions \cite{d2018automated,schwaner2021autonomous} or sampled-based planning methods \cite{lu2020dual,ozguner2021visually}. 
Recent methods learns controllers via RL without the need for planning \cite{xu2021surrol,chiu2021bimanual,bendikas2023learning,long2023human,huang2023demonstration}.
Xu et al. formulated the surgical grasping with the goal-based Markov Decision Process.
Their policy uses the track pose of target objects and robot proprioceptive measurements as input and is learned with hindsight experience replay (HER) \cite{andrychowicz2017hindsight} efficiently given sparse rewards and a set of demonstration trajectories \cite{xu2021surrol}. 
Bendikas et al. decompose the grasping task into subtasks to reduce the difficulty in RL learning \cite{bendikas2023learning}.
However, needle pose tracking is extremely challenging due to object occlusion, the thin shape of needles, and complex backgrounds \cite{li2024monocular}.
Researchers aim to alleviate the issues of needle pose estimation in surgical grasping through various strategies.
Sen et al. \cite{sen2016automating} designed customized grippers that constrain the needle’s orientation during closure.
Sudaresan et al. segment image masks of needles to reduce background noise \cite{sundaresan2019automated}.
Wilcox et al. improve visibility by adjusting the needle’s position to a predefined pose \cite{wilcox2022learning}.
Other surgical objects, including threads \cite{joglekar2023suture,lu2019surgical}, tissue fragments \cite{kehoe2014autonomous,seita2018fast,fan2024learn} and blocks \cite{hwang2022automating,xu2021surrol} are also explored.
The Cartesian poses of these objects are tracked, and robots are controlled to grasp them based on the tracked poses, either through planned trajectories \cite{kehoe2014autonomous,seita2018fast,hwang2022automating} or learned controllers \cite{fan2024learn,xu2021surrol}.
Some works develop general frameworks to learn controllers that grasp diverse objects \cite{xu2021surrol,long2023human,huang2023demonstration}. 
Yet they are required to learn separate controllers for their corresponding target objects.
In contrast, our approach learns a single visuomotor controller that is capable of handling diverse grasping tasks.
The aforementioned previous methods rely on high precision of pose estimation for target objects, which is difficult to achieve in real-world surgical settings due to imprecise robot proprioceptive measurements, complex visual disturbances, and the millimeter-scale size of surgical objects.
Moreover, none of these studies have validated their feasibility in real surgical environments.
Finally, these approaches struggle to generalize to unseen objects due to their reliance on predefined geometric models or shape assumptions.

A few studies have explored visuomotor learning for surgical grasping as a potential alternative to overcome the limitations of pose-tracking methods \cite{scheikl2022sim,kim2024surgical,huang2024visuomotor}.
Scheikl et al. \cite{scheikl2022sim} train a visuomotor policy in simulation for tissue retraction using Proximal Policy Optimization (PPO) \cite{schulman2017proximal}, a model-free RL method.
Sim-to-real visual discrepancy is minimized with a visual domain adaptation technique, unpaired image-to-image translation models.
When deployed on real robots, the learned policy achieved a $50\%$ success rate.
However, the poses of both the gripper and the target object's grasping point are still required to be tracked to construct their reward function, which hardly alleviates the limitations of pose tracking. 
In contrast, we do not rely on pose tracking on task objects. 
Furthermore, while they assume a fixed initial pose of the target object, we focus on a more challenging surgical grasping problem where the object's initial pose is arbitrarily positioned.
Learning visuomotor policy with imitation learning(IL) is also explored \cite{huang2024visuomotor,kim2024surgical}. 
Huang et al. are the first to train real-time and closed-loop control in an end-to-end IL paradigm for surgical tasks \cite{huang2024visuomotor}.
The trained Visuomotor policy, running at 30Hz control frequency, achieves $75-100\%$ success rate in simulation for needle grasping and peg transfer.
However, it remains unclear whether their method can bridge the sim-to-real gap, as it has not been evaluated on real robots.
Moreover, they rely on accurate measurement of joint positions as their policy's observation, which is challenging for surgical robots like da Vinci Surgical System.
Kim et al. relieve the assumption of accurate robot proprioception by using approximate kinematics data without further kinematics corrections \cite{kim2024surgical}.
They successfully showcase their trained visuomotor policy can perform surgical tasks (tissue lift, needle grasping and knot tying) on real robots with nearly $100\%$ success rate and handle unseen clinically relevant backgrounds.
However, their method are limited to the following aspect: First, their controller relies on visual observations from additional wrist cameras, which is challenging because it requires a larger incision port due to the larger instrument size, increases the expense of surgery, and suffers from more visual disturbances, such as blood occlusion, compared to the endoscopic camera, since it is fixed on the moving instrument.
Second, they require to collect 200-500 human demonstrations for each surgical task on real robots, which require extensive human efforts for data collection.
Finally, they only showcase quantitative studies on handling unseen backgrounds that are similar to those encountered during training.
It is unclear if their controller can handle unseen OOD backgrounds with high diversity.
Compared to previous works on visuomotor learning for surgical robots, we train a visuomotor policy via a model-based RL method in simulation and directly transfer it to real robots without the need of collecting demonstration on real robots.
In addition, our method does not rely on either robot proprioceptive measurement or tracking target pose.
We also showcase our controller can handle a high diversity of OOD scenarios and disturbances.

\subsection{Visuomotor Learning for Robot Grasping}
Visuomotor learning can be classified into two lines of research based on the action types: grasp pose detection and continuous action learning.
The first line defines the grasping problem as predicting multiple poses within a given scene, typically in Cartesian space, to enable the robot to successfully grasp objects by positioning its end-effector accordingly.
Zeng et al. propose an affordance-based perception framework for object-agnostic grasping, which plans four primitive grasping actions and selects the most suitable one based on dense affordance probability maps \cite{zeng2022robotic}.
Their method also enables object recognition for both known and novel items using only product images, without the need for additional data collection or retraining.
Mousavian et al. introduce a variational autoencoder (VAE) for mapping partial point clouds to grasp predictions and a grasp evaluator network for assessing grasp quality, improving overall grasp planning \cite{mousavian20196}.
However, these two methods rely on open-loop grasping and cannot adjust grasping poses dynamically during execution.
Fang et al. develop the first unified closed-loop system for fast and accurate grasping, incorporating object center-of-gravity awareness and a novel generation-association method for dynamic 7-DoF grasp prediction \cite{fang2023anygrasp}.
Although their method mitigates the limitations of open-loop approaches, their controller struggles to handle occlusions caused by the robot.
A key limitation of grasp-pose-prediction approaches is their dependence on precise robot kinematics to position the gripper accurately for grasping.
These methods face challenges in handling the substantial kinematic errors often encountered in surgical robots.

Another line of visuomotor learning, continuous action learning, directly maps the observation to a continuous action---including motor torques, joint position, and relative gripper pose.
Some works investigate model-free methods for visuomotor learning, where the visuomotor controller is trained without explicit dynamic models of environments.
Song et al. introduce a visual 6-DoF closed-loop grasping algorithm using action-view rendering and collect human grasping demonstrations in diverse environments with a new low-cost hardware interface \cite{song2020grasping}. 
Wang et al. present the Goal-Auxiliary Deep Deterministic Policy Gradient (GA-DDPG) algorithm, which combines imitation and reinforcement learning for 6D robotic grasping, and demonstrate its effectiveness in improving grasping performance within a tabletop system \cite{wang2022goal}.
Chebotar et al. introduce a goal-conditioned Q-learning approach with hindsight relabeling to train robots in a challenging offline setting, enabling them to learn diverse skills from high-dimensional camera images and generalize to novel scenes and objects \cite{chebotar2021actionable}.
Lee et al. develop a vision-based policy through reinforcement learning that enables stacking various non-cuboid objects with diverse strategies, achieving these results without human demonstrations \cite{lee2021beyond}.
Some works explore leveraging task priors to improve visuomotor learning \cite{walke2023don,rusu2017sim,zeng2021transporter}. 
Walke et al. demonstrate that integrating prior data into a reinforcement learning system improves sample efficiency \cite{walke2023don}.
Rusu et al. initialize a new network with lateral, nonlinear connections to a simulation-trained network and train it on a similar task on a real robot, finding that the inductive bias from the simulation significantly accelerates learning on the real robot \cite{rusu2017sim}.
Zeng et.al propose the Transporter Network, a model that leverages exploiting spatial symmetries as a task prior to infer spatial displacements from visual input for robot actions, significantly outperforming alternative methods in learning vision-based manipulation tasks \cite{zeng2021transporter}.
Some researchers investigate the benefits of using large datasets collected from real robots\cite{kalashnikov2018scalable,levine2018learning,kalashnikov2021mt}.
Kalashnikov et al. introduce QT-Opt, a scalable self-supervised vision-based reinforcement learning framework that leverages 580k real-world grasp attempts to train a deep Q-function, achieving strong grasp performance on unseen objects, while automatically learning regrasping strategies, probing objects for optimal grasps, repositioning objects, and handling disturbances and non-prehensile manipulations \cite{kalashnikov2018scalable}.
Sergey et al. propose a method for continuous visual servoing in robotic grasping from monocular cameras, using a Convolutional Neural Network (CNN) for predicting grasp outcomes and a large-scale data collection framework with 800-900k grasp attempts. S
They further propose a method for continuous visual servoing in robotic grasping from monocular cameras, using a novel CNN for predicting grasp outcomes and a large-scale data collection framework with 800-900k grasp attempts \cite{levine2018learning}.
Kalashnikov et al. propose MT-Opt, a scalable multi-task reinforcement learning method, and a system for simultaneously collecting data across tasks, enabling robots to share experience and improve performance through user-provided task examples \cite{kalashnikov2021mt}.

Although model-free methods can train visuomotor policies with strong performance in robotic grasping, they struggle with low data efficiency and generalization compared to model-based approaches \cite{asadi2015strengths}.
This has led to growing interest in model-based methods for their potential to improve both efficiency and generalization \cite{moerland2023model}.
Some works investigate a visual MPC framework, Visual Foresight \cite{ebert2018visual,dasari2019robonet,xie2019improvisation}.
Ebert et al. are the first to present visual foresight for deep reinforcement learning that leverages sensory prediction models to learn behaviors in diverse environments while effectively predicting pixel-level observations under occlusions and novel objects \cite{ebert2018visual}.
Dasari et al. explore visual foresight in the context of multi-robot, multi-domain tasks \cite{dasari2019robonet}. 
Xie et al. combine imitation learning and visual foresight, integrating imitation-driven models into data collection and planning \cite{xie2019improvisation}. 
Recent advances in world models \cite{hafner2019dream,hafner2020mastering} draw interest in the field of robot grasping.
Compared to visual foresight, world models predict environment dynamics under learned latent space instead of image space, facilitating faster computation for long-horizon tasks. 
The visuomotor policies, trained by planning \cite{mendonca2023structured,mendonca2023alan}, imitation learning \cite{mandi2022cacti,lancaster2023modem}, and RL \cite{wu2023daydreamer,seo2023multi} based on the learned world models, are applied to visual manipulation tasks in real robots.
Wu et al. deploy world models in robot grasping, learning a visuomotor policy directly on a real robot from scratch \cite{wu2023daydreamer}.
Mendonça et al. leverage internet videos of human interactions to pre-train world models \cite{mendonca2023structured} and use these models for exploration \cite{mendonca2023alan}. 
Mandi et al. propose CACTI, a framework for scaling robot learning in kitchen manipulation that integrates data collection, augmentation, visual representation learning, and imitation policy training, using generative models and pretrained visual representations for improved efficiency \cite{mandi2022cacti}.
Lancaster et al. demonstrate the first successful real-world application of demonstration-augmented visual model-based RL, addressing unsafe exploration and over-optimism \cite{lancaster2023modem}.
Seo et al. present the Multi-View Masked World Model, a reinforcement learning framework that learns a world model using a multi-view masked autoencoder, adapting to various viewpoints \cite{seo2023multi}.

Yet, none of the previous works on visuomotor learning for robotic grasping have demonstrated the ability to handle millimeter-scale objects---for example, a surgical needle, which has a width of approximately 1 mm.
Compared to their works, our learned visuomotor policy grasps millimeter surgical objects with strong performance. 
For system robustness, background noise is filtered out by a learned mask in \cite{mendonca2023alan}, and viewpoint randomization is applied to adapt viewpoint variations and disturbance \cite{seo2023multi}.
However, in surgical grasping, the mask area of task objects in the observed images is much smaller than that in \cite{mendonca2023alan}.
Furthermore, we apply more comprehensive domain randomization, including camera pose, target disturbance, gripper disturbance, and image noise, for visuomotor learning compared to \cite{seo2023multi}.
James et.al. are the first to use a voxel representation for vision-based reinforcement learning for 6D robot manipulation \cite{james2022coarse}.
However, the effectiveness of voxel representation for surgical grasping is still unclear.
Compared to their work, we use voxel representation for orthogonal projection, where the resulting masks are equivariant with respect to the object's translation.

\section{Problem Formulation}

\subsection{Problem Setup}
\label{sec: problem setup}
Our goal is to control a robotic arm with an arbitrary gripper to grasp an arbitrary target object within a workspace via a learned visuomotor policy $\pi$ that maps temporal observation to an action command. 
The size of the gripper and the target object is within a ball with a diameter of 5cm.
The initial poses of the target object and the robot gripper are set randomly within the workspace. 
Visual observations are captured using an endoscopic stereo camera. 
At timestep $t$, the camera captures a pair of RGB images---left ($I_l$) and right ($I_r$) images---each with a resolution of $600 \times 600 \times 3$. 
A normalized command $a \in \mathbb{R}^{5}$, whose elements are in the range of $[-1,1]$, controls the actuation of the robotic gripper.
The first four elements of the action vector control the Cartesian translation of the gripper along the $x$, $y$, and $z$ axes, as well as its Cartesian orientation along the $z$-axis.
These elements represent delta action commands, constrained by the maximum translational and rotational magnitudes, $\Delta^{xyz}$ and $\Delta^{\theta}$, respectively.
The delta action commands for the gripper's translation and rotation are preferred over absolute commands due to their robustness against the imprecise kinematics of surgical robots, as highlighted in \cite{kim2024surgical}.
The last element of the action vector controls a binary open/close state of the gripper's jaw, where a positive value actuates the gripper to open, and a non-negative value actuates it to close.
The task is considered successful if the gripper grasps the target object.
When the gripper's desired position lies outside the defined workspace, the system actuates the gripper to the nearest feasible point within the workspace boundary.

\subsection{Assumptions} 

We assume that i) the joint positions of the robot arm remain within the robot's joint limits throughout the task;
ii) both the target objects and the gripper may be partially occluded, but are never fully occluded;
and iii) the poses of the target objects and the gripper cannot be tracked explicitly using the robot’s kinematics or through camera observations.
 
\section{Method}

We present our framework, GASv2, for visuomotor learning in surgical grasping tasks.
We begin by formulating the problem, followed by introducing our network architecture and training pipeline.
Next, we elaborate on the visual perception of visuomotor learning in surgical grasping.
Finally, we present our hybrid control architecture and training environments.

\subsection{Formulation}
We formulate the control problem of general surgical grasping as a discrete-time partially observable Markov decision process (POMDP).
POMDP consists of a 7-tuple $(S, A, R, T, O,\gamma, \Omega)$, where $S$ is a set of partially observable states; 
$A$ is a set of discrete actions; 
$R(s, a): S\times A \to \mathbb{R}$ is a reward function; 
$T$ is a set of conditional transition probabilities between states; 
$O$ is a set of conditional observation probabilities; 
$\gamma \in [0,1]$ is the discount factor; and
$o \in \Omega$ is the observation.
We aim to learn the control policy that maximizes its expected future discounted reward
$\mathbb{E}_{\pi}[ \sum_{i=t}^{H} \gamma^{i-t}r_{i}]$, where $r_i$ is the reward at time $i$, and $H$ is the number of timestep in a episode.

\subsubsection{Termination and Reward Design}
\label{sec: Termination and Reward Design}
Constructing a shaped reward function $R(s, a)$, known as \textit{dense reward}, is challenging for general surgical grasping since it necessitates expert-level domain knowledge and cumbersome engineering \cite{lin2023end}.
Furthermore, dense rewards in previous works of surgical grasping require tracking object poses explicitly \cite{scheikl2022sim}, which is not available in our assumption. 

We use \textit{sparse rewards} to alleviate such limitations, where reward signals are discrete and finite.
In particular, four task-level states, directly determining both rewards and terminations, are defined as
\begin{itemize}
    \item \textbf{Successful Termination}: When the gripper is closed and the target object is successfully grasped, a constant positive reward of $1$ is given and the task is terminated.
    \item \textbf{Failed Termination}: If i) the gripper closes without a successful grasp or ii) the current timestep reaches the maximum horizon $H_{\text{max}}$, a constant negative reward of $-0.1$ is assigned and the episode is terminated.

    \item \textbf{Abnormal Progress}: When any of the following two abnormalities occur: i) the gripper's desired pose exceeds the workspace; or ii) the gripper's vertical height falls below a specified threshold relative to the target object (as detailed in Sec. \ref{sec: Safe Control for Handling Platforms with Height Variance}), a constant negative reward $-0.01$ is provided to discourage the occurrence of these events.
    In this task-level state, the task is not terminated.
    \item \textbf{Normal Progress}: In this state, the task proceeds normally.
    A constant negative constant reward of $-0.001$ is given to speed up the grasping.
    The task is not terminated in such a task-level state.
\end{itemize}
Since these task-level states are finite and their transitions are fully defined, a finite-state machine (FSM) is used for the transitions of the states.
In particular, each task-level state is defined as a corresponding FSM's state. 
The FSM's transition is fully determined by the transition rules of task-level states.
Notably, since the task's termination is fully controlled by the FSM, the action of our controller can not directly control the termination of the task.

\subsubsection{Observation and Action Spaces}
\label{sec:obs_and_act_spaces}
The observation of our visuomotor policy consists of two parts: a compact image and three robot system states.
The size of the compact image is $64\times64\times3$.
The robot system states are three real variables $s^{sys}_1,s^{sys}_2,s^{sys}_3\in  \mathbb{R}$, corresponding to i) FSM state, ii) gripper state, a nd iii) controller state, respectively.
In particular, $s^{sys}_1$ encodes the ID of our task-level states in our FSM;
$s^{sys}_2$ encodes the joint position of the gripper's jaw;
and $s^{sys}_3$ encodes ID of sub-policies in our hybrid control architecture, which will be introduced in the latter Sec. \ref{sec:control_phase}. 
Systems states are normalized to the range of $[0, 1]$.

To alleviate the difficulty of RL training, discrete actions are used for our visuomotor policy.
The discrete action space, containing 9 discrete actions, is denotes as $A_{\text{d}}=\{\pm a^x_{d},\pm a^y_{d},\pm a^z_{d},\pm a^{\theta}_{d},a^{\lambda}_{d}\}$, where $a^x_{d}=[1,0,0,0,0]$, $a^y_{d}=[0,1,0,0,0]$, $ a^z_{d}=[0,0,1,0,0]$, $a^{\theta}_{d}=[0,0,0,1,0]$, and $a^{\lambda}_{d}$ correspond to the $x,y,z$ translation action, rotation action and jaw toggle action, respectively. If the jaw toggle $a^{\lambda}_{d}$ is activated, the jaw will move based on the current jaw state. If the jaws are opened, it closes the jaw with a robot command $[0,0,0,0, -1]$. Otherwise, if the jaws are closed, it opens the jaw with a robot command $[0,0,0,0, 1]$.

\subsection{Network Architecture and Training}
\label{sec:Network Architecture and Training}
Our visuomotor policy is learned with a world-model-based algorithm, DreamerV2 \cite{hafner2020mastering}.
DreamerV2 builds a generative world model that captures the dynamics of POMDP. 
The world model is learned with past ground-truth experiences sampled from a replay buffer.
A visuomotor policy is learned from generative transitions from the world model with an actor-critic algorithm. 
The robot agent generates ground-truth trajectories by interacting with the environment.
The resulting ground-truth transitions are then stored in the replay buffer for off-policy learning.
Fig. \ref{fig:gasv2_overview}d illustrates the pipeline of DreamerV2.

For world models, observation is encoded to a stochastic latent state $z$ through a Variational Auto-Encoder (VAE) \cite{kingma2013auto}.
Then, the sequences of the latent states are predicted by Recurrent State-Space Model (RSSM) \cite{hafner2019planet}, a sequence model with a deterministic recurrent state $h$.
Finally, the reward, the discount factor, and reconstructive observation $\hat{o}$ are predicted based on the encoded hidden state, which is formed by the concatenation of these states as $s=[h, z]$.
A prior hidden without accessing current observation is given by $\hat{s}=[h, \hat{z}]$, where $\hat{z}_t$ is the estimated stochastic hidden state predicted by the transition predictor of RSSM $\hat{z}\sim p_{\phi}(\hat{z}|h)$.
The control policy is learned by the actor-critic mechanism, where an actor  $p_{\psi}(\hat{a} | \hat{z})$ and a critic $v_{\xi}(\hat{z})$ are used to predict the action and the value, respectively.
The actor network learns a distribution of actions for each latent model state, aiming to maximize the cumulative future task rewards.
Meanwhile, the critic network is trained using temporal difference learning to predict the total future task rewards \cite{sutton2018reinforcement}.

To learn the world model and the policy, Adam optimizer is used as a common practice \cite{kingma2014adam}.
Both world models and control policy are optimized jointly.
To accelerate learning convergence, we prefill the buffer with transitions from rollout trajectories generated using a scripted demonstration policy. 
Additionally, we include transitions from a random policy to ensure uniform sampling of the environment's dynamics.
Details of the DreamerV2 implementation are available in \cite{hafner2020mastering}, while the modified hyperparameters used in this work are summarized in Table \ref{table:hyperparam}.

\subsection{Visual Perception}

Training a visuomotor controller with strong performance heavily relies on the choice of the controller's visual input.
As defined in Section~\ref{sec:obs_and_act_spaces}, we aim to encode the stereo images, $I_l$ and $I_r$, each with a resolution of $600\times600\times3$, into a compact image of resolution $64\times64\times3$ for the visuomotor policy's input.
The original implementation of DreamerV2 is downsampling stereo images to obtain the input image \cite{wu2023daydreamer}. 
However, as shown in our ablation experiments in Section~\ref{sec:exp_ablation}, downsampling images leads to a degradation in policy performance.
This performance drop is caused by significant image quality loss during downsampling, which limits the policy’s ability to encode the high-resolution spatial information of task-relevant objects.
In addition, inferring the spatial information of task-relevant objects from implicit image representations--- two stereo RGB images---is significantly more challenging for training image encoders in visuomotor policies, compared to using explicit spatial representations like depth images or point clouds.
To address these limitations, we propose a surgical perception pipeline for our visuomotor controller, which significantly enhances its efficiency.
Fig. \ref{fig:gasv2_overview}b shows the overview of our perception pipeline.
We first estimate the depth and object masks based on the stereo images.
Then we re-project the depth and mask to images using orthogonal projection. 
Finally, we use a novel image representation for the visuomotor's input image.

\begin{figure*}[!tbp]
  \centering
  \includegraphics[width=1.0\hsize]{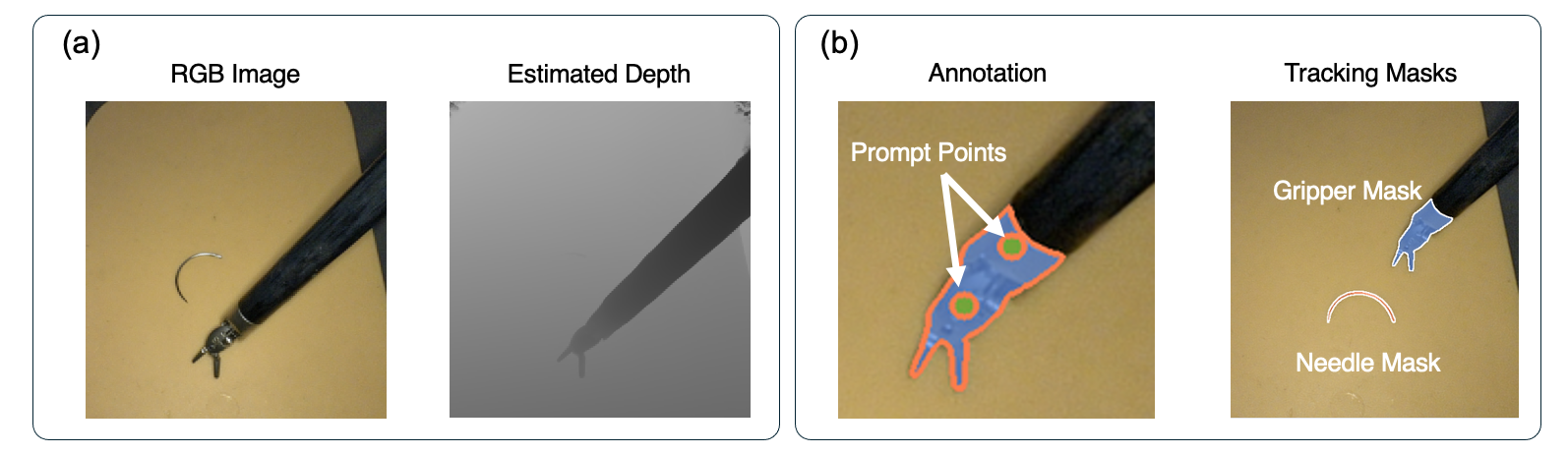}
  \caption{Our depth estimation and visual segmentation. (a) An RGB image from the left stereo camera, along with its estimated depth, are shown. (b) Two masks representing the gripper and target object in our video object segmentation are displayed. On the left, points are prompted to annotate the gripper. On the right, these masks are tracked based on temporal left stereo RGB images. }
  \vspace{-0.45cm}
 \label{fig:segment_and_depth_estimation}
\end{figure*}

\subsubsection{Depth Estimation with Stereo Matching Network}
\label{sec:depth estimation}

Since implicitly inferring the spatial information of objects using the policy's image encoder is challenging, we explicitly estimate the depth of visual observation.
To this end, we first calibrate the intrinsic parameters of the stereo camera using a toolbox, OpenCV \cite{opencv_library}.
Based on the calibrated parameters, stereo images are processed by center cropping and rectification, resulting in rectified images $I'_l, I'_r \in \mathbb{R}^{600\times600\times3}$. 
The resultant rectified images are further used to infer depth information via stereo matching \cite{hamid2022stereo}.
For each pixel with image coordinates $p=(p_x,p_y)$ in the left image, a corresponding pixel with the coordinate $(p_x-w,p_y)$ exists in the right image, where $w$ denotes the disparity for the $p$ coordinate of the left image.
The pixel correspondence and disparity are estimated by a state-of-the-art (SOTA) pre-trained stereo matching network, IGEV \cite{xu2023iterative}, which uses iterative geometry encoding to refine depth estimation.
This approach enhances robustness and generalization of depth estimation, particularly in textureless or complex regions encountered during surgery.
The depth $d_{p}$ for each pixel with coordinates $p$ in the left image is calculated as
\begin{equation}
    d_{p} = \frac{C_f \cdot C_b}{w},
\end{equation}
where $C_f$ is the camera's focal length, and $C_b$ is the baseline distance between the two camera centers. 
Both $C_f$ and $C_b$ can be derived from the calibrated intrinsic parameters.
The correspondence coordinates of the resultant depth image, $I^d=\{d_{p}\}^{600\times600}$, are aligned with those of the left rectified stereo image.
Fig. \ref{fig:segment_and_depth_estimation}a shows a left rectified stereo image and its corresponding depth image.

\subsubsection{Semi-Supervised Video Object Segmentation}
\label{sec:segmetation}
A key challenge in visuomotor learning is reducing background interference while accurately identifying the object within the spatial context.
As motivated before, this challenge is further exacerbated in the context of surgical grasping, where the region of interest---i.e., the gripper and target object---comprises only $1\%$ of the entire image.

To address the challenge, we explicitly infer the object's pixel-wise mask. The segmentation task is formulated as the problem of video object segmentation (VOS): given a sequence of observed temporal RGB images, we compute a set of binary masks  $M =\{m_i \in\mathbb{R}^{600\times600}\}_{i=1}^K$ for $K$ task objects. 
Object masks are inferred from the left rectified RGB image. We prefer using the left rectified RGB image for segmentation over the estimated depth image, as the latter is generally noisier and less reliable than stereo RGB images.
Each element in the masks is assigned a corresponding pixel class label.
In surgical grasping, we assign two class labels ($K=3$)---gripper,  target object, and background. Fig. \ref{fig:segment_and_depth_estimation}b shows masks of both the gripper and the target object.

Tracking object masks in surgical scenes poses several challenges, including target deformation, camera motion \cite{yang2023track}, motion blur \cite{allan20192017}, illumination variations such as shadows or specular reflections, as well as occlusions from factors like blood or lens fogging \cite{shvets2018automatic}.
Moreover, training data-driven VOS methods relies on large manually annotated datasets, which require significant human effort and are both time-consuming and tedious \cite{garcia2021image}.

To overcome these challenges, we utilize a semi-supervised VOS method, Track Anything Model (TAM) \cite{yang2023track}, combining two SOTA pre-trained models, Segment-Anything Model (SAM) \cite{kirillov2023segment} and XMem \cite{cheng2022xmem}.
Compared to methods that train mask detectors from scratch \cite{redmon2016you}, TAM eliminates the need for labeling datasets or model training for segmentation.
TAM enables object segmentation in video streams with minimal annotation effort, requiring only a few clicks for annotation.
For each rollout of our visuomotor policy, we infer the object masks using TAM in a two-stage process:

\begin{itemize} 
\item \textbf{Annotation Stage}: In this stage, a user annotates the masks for the first frame of the temporal RGB images in a rollout.
With SAM, fine masks can be created via a few point prompts---approximately two clicks per object mask.
Fig. \ref{fig:segment_and_depth_estimation}b shows the point prompt.
The annotation is then saved for tracking. 

\item \textbf{Online Tracking Stage}: Given the annotation, XMem automatically tracks object masks for the remaining frames by leveraging both temporal and spatial correspondence of streaming input images. Fig. \ref{fig:segment_and_depth_estimation}b shows the tracked masks of the gripper and the target object. 
\end{itemize}
We were surprised to find that annotated masks, when applied to contexts with similar objects and backgrounds, could be reused in our ex vivo experiments (see Sec. \ref{sec:Ex-vivo Animal Study}), significantly reducing the annotation effort.

After obtaining segmentation masks, we segment the depth image $I_d$ using the predicted masks.
Since the correspondence between coordinates in the left rectified stereo image and the estimated depth image is aligned, the obtained masks can be directly applied to segment the depth image.
A set of depth segments can be formulated as $\bar{I}^d = \{\bar{I}^d_{i}:\bar{I}_i^d=I^d\odot m_i\}_{i=1}^K$, where $\odot$ is the operator of element-wise multiplication.

\subsubsection{Image Reprojection with Voxel-Based Orthogonal Projection}

\begin{figure*}[!tbp]
  \centering
  \includegraphics[width=1.0\hsize]{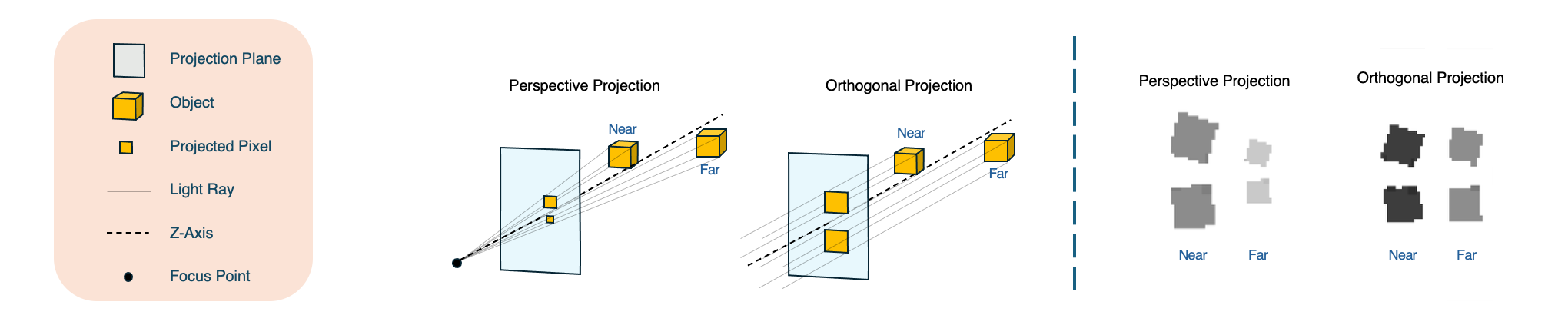}
  \caption{On the left, we illustrate the principles of image formation for both perspective and orthographic projections. Two camera-to-object distances---one far and one near---are compared. In perspective projection, the image is formed by radiant beams converging at a focus point, while in orthographic projection, the image is formed by parallel beams.
On the right, we show depth images captured at two different camera-to-object distances using both projection types.
While perspective projection does not preserve consistent shape scaling, orthographic projection maintains shape equivariance.}
  \vspace{-0.45cm}
 \label{fig:orthogonal_projection}
\end{figure*}

Previous visuomotor learning methods typically use perspective-projected images as the controller's visual input. In perspective projection, image pixels are generated by radial rays that converge at the camera's focal point. Fig. \ref{fig:orthogonal_projection} illustrates the principle of perspective projection. In this paper, the stereo RGB images $I_l, I_r$, the estimated depth $I^d$, and the inferred masks $M$ all adhere to the principle of perspective projection.

However, images with perspective projection fail to maintain \textit{equivariance} for the image encoder based on convolutional neural networks (CNNs).
Equivariance refers to the property where translating an object in the workspace and then encoding the resulting image should yield the same feature maps as encoding the original image and then translating the feature maps.
Convolutional neural networks are inherently equivariant to translations along the $x$ and $y$ axes, where object movement results in corresponding pixel shifts in the image.
However, they are not equivariant to translations along the $z$ axis, which change the object’s scale in the image, breaking equivariance \cite{cohen2016group}.
Fig. \ref{fig:orthogonal_projection} demonstrates the scaling effects of depth images as the object translates in the $z$ axis of the camera's frame.

To preserve equivariance in image representations, we employ a voxel-based orthogonal projection method \cite{lin2025Multi} to reproject both depth images and segmentation masks into an equivariant form with respect to the translation of task objects.
The reprojection pipeline consists of three main steps: inferring point clouds, voxelization, and orthogonal projection.

First, we infer the point clouds from our estimated depth image.
For each depth pixel $\bar{d}_{p,i}$ in segmented depth $\bar{I}_{i}^d$ with image coordinates $p=(p_x,p_y)$ of task object $i$, we calculate the corresponding 3D Cartesian point $q_{p,i}$ based on the camera's intrinsic and extrinsic parameters, following the camera pinhole model \cite{szeliski2022computer}, which is given by:
\begin{equation}
    q_{p,i} = R_{cam}[\alpha_{cam} p_x \quad \alpha_{cam} p_y \quad \bar{d}_{p,i}]^{T} + l_{cam},
\end{equation}
where $\alpha_{cam}$ is a scaling factor related to the camera's intrinsic parameters, and $R_{cam}\in\mathbb{R}^{3\times3}$ and $l_{cam} \in\mathbb{R}^{3}$ represent the rotation matrix and translation vector of the camera's reference frame, respectively, as specified by its extrinsic parameters.
The corresponding point cloud segment for object $i$ can be represented as $Q_i=\{q_{p,i}\}$.

After we infer the point cloud segments, we apply voxelization to the inferred point clouds . 
In particular, we evenly rasterize the 3D Cartesian workspace into $N_{voxel}\times N_{voxel}\times N_{voxel}$ identical cuboids along $x$, $y$, and $z$ axes, where $N_{voxel}\in \mathbb{R}^+$ is the number of voxel along an axis \cite{xu2021voxel}.
These cuboids, called voxels, are indexed as $B=\{b_{x,y,z}\}$, where $b_{x,y,z}$ represent a voxel with index coordinates $(x,y,z)$ along $x,y$ and $z$ axes. 
Afterward, we identify the voxels that contain at least one point from the point cloud $Q_i$ for object $i$.
The identified voxels are defined as $B_i=\{b^i_{x,y,z}: b_{x,y,z} \text{ contains points in } Q_i \}\subseteq B$.
To reduce the noise caused by errors in our depth estimation and segmentation, we filter out the elements of noisy voxels in the identified set $B_i$ using a sphere-based neighbor method \cite{zhou2018open3d}.
We also compute a voxel-based, low-dimensional observation signal consisting of centroid coordinates. This signal is used for image representation (Section \ref{sec:image_representation}), phase classification (Section \ref{sec:control_phase}), as input to a PID sub-policy (Section \ref{sec:control_subpolicy}) and safe trajectory dealing with platforms with height variance (Section \ref{sec: Safe Control for Handling Platforms with Height Variance}).
In particular, for task object $i$, the centroids of its identified voxels
$(c_x^i,c^i_y,c^i_z)$ are
\begin{equation}
\label{equ:voxel_centroids}
\begin{aligned}
    c_x^i &= \frac{\sum_{b^i_{x,y,z}\in B_i} x}{N_{voxel}} ,\\
     c_y^i &= \frac{\sum_{b^i_{x,y,z}\in B_i} y}{N_{voxel}} , \\
      c_z^i &= \frac{\sum_{b^i_{x,y,z}\in B_i}z}{N_{voxel}}.
\end{aligned}
\end{equation}

Finally, we orthogonally project the identified voxels $B_i$ for task object $i$ along the $z$ axis of the voxels.
The projected light rays are parallel with the $z$-axis as shown in Fig. \ref{fig:orthogonal_projection}.
Let $B_{\bar{x},\bar{y},i} = \{ b_{x,y,z}\in B_i: x=\bar{x}, y=\bar{y}\}$ be a subset of $B_i$ contains the object $i$'s voxels at the coordinates of $\bar{x}$ and $\bar{y}$.
We obtain a projected depth image $I^{proj}_i=\{d_{\bar{x},\bar{y},i}\}^{N_{voxel}\times N_{voxel}}$ for object $i$, where each pixel $d_{\bar{x},\bar{y},i}$ with image coordinate $(\bar{x},\bar{y})$ for object $i$ is given by 
\begin{equation}
    d_{\bar{x},\bar{y},i} = \begin{cases} \min\limits_{b_{x,y,z} \in \bar{B}_{\bar{x},\bar{y}, i}} z &   \text{if }\bar{B}_{\bar{x},\bar{y}, i} \neq \emptyset \\
    0 & \text{otherwise}
    \end{cases}.
\end{equation}
Likewise, we obtain a projected mask $m^{proj}_i=\{m_{\bar{x},\bar{y},i}\}^{N_{voxel}\times N_{voxel}}$ for object $i$, where each pixel $m_{\bar{x},\bar{y},i}$ with image coordinate $(\bar{x},\bar{y})$ for object $i$ is given by 
\begin{equation}
    m_{\bar{x},\bar{y},i} = \begin{cases} 1 &  \text{if }\bar{B}_{\bar{x},\bar{y}, i} \neq \emptyset\\
    0 & \text{otherwise}
    \end{cases}.
\end{equation}


\subsubsection{Image Representation with Dynamic Spotlight Adaptation}
\label{sec:image_representation}

As motivated before, high-resolution visual observations must be encoded into a compact image representation for the visuomotor policy's input, due to limitations in model capacity and computational cost.
For DreamerV2, its image encoder, which uses a Variational Autoencoder (VAE), requires input images with a compact resolution of  $64\times64\times3$, as VAEs are notoriously difficult to scale for high-dimensional images \cite{kingma2013auto, hafner2020mastering}.
Furthermore, increasing visual resolution inevitably requires a larger model capacity, which increases the computational complexity of training world-model-based algorithms and limits the benefits of training on a single GPU without relying on a GPU cluster.

We propose a novel image representation for visuomotor policy in surgical grasping, Dynamic Spotlight Adaptation (DSA), which effectively satisfies the constraints of compact visual input while maintaining high resolution in critical local regions.
Specifically, DSA channel-wise encodes the orthogonal projected depths and masks as well as robot system states into an image with a lower resolution of $64\times 64\times3$.
The encoded image consists of three layers (channels): the first encodes the orthogonal depths, the second encodes the orthogonal mask, and the third encodes the system states.

We start with obtaining the first layer for the projected depths.
First, the pixel values in the projected depth $I_i^{proj}$, ranging from $1$ to $N_{voxel}$, are truncated to a smaller range centered around the $z$-centroid coordinate $c_z^i$ of the gripper's identified voxels. 
The range width is $N_{zoom}\in(0,N_{voxel}]$.
Afterward, we scale the truncated depth to a depth image with a standard $8$-bit unsigned integer (UInt8) range of $[0,255]$.
Finally, we zoom in the vicinity of the gripper's pixels with a box mask $m_{zoom}\in\mathbb{R}^{N_{zoom}\times N_{zoom}}$ to obtain a local-level image.
Since the gripper's pixels on the image will shift as the gripper's location changes, we need to dynamically adjust the coordinates of the zoom-in mask to ensure that the gripper's pixels remain within view in the local-level image.
To this end, the mask centroids are aligned with the centroids of the gripper’s identified voxels, computed as $(c_x^i \cdot N_{voxel},\, c_y^i \cdot N_{voxel})$, where $\cdot$ denotes the multiplication operator, and $c_x^i$ and $c_y^i$ are the $x$ and $y$ centroids of the gripper’s voxels defined in (\ref{equ:voxel_centroids}).
The zoom-in process can be viewed as first segmenting the image using a given mask, then resizing it to a target size. Accordingly, we first segment the zoom-in mask $m_{zoom}$ and then resize it to $64\times64$.
The first channel layer of the DSA image $I_1^{DSA}$ is obtained by summing the resultant images as
\begin{equation}
    I_1^{DSA} = \sum_{i=1}^K\mathcal{R}(\Phi(\mathcal{S}(\mathcal{T}(I_i^{proj})))),
\end{equation}
where $\mathcal{T}()$, $\mathcal{S}()$, $\Phi()$ and $\mathcal{R}()$ are the truncation, scaling, segmentation, and resize functions, respectively.

Next, we encode the projected mask $m_i^{proj}$ for the second layer of the DSA image.
In particular, for each task object $i$, it corresponding projected mask is multiplied with an encoding scalar $e_i$  with the UInt8 range of $[0,255]$.
After the multiplication, we apply the same zoom-in process as that of the first layer.
We zoom-in within the area of the zoom-in mask $m_{zoom}$, where we first segment the encoding mask with the zoom-in mask $m_{zoom}$ and then resize it to $64\times64$.
The second layer of the DSA image $I_2^{DSA}$ is obtained by summing the resultant images as
\begin{equation}
I_2^{DSA} =\sum_{i=1}^K \mathcal{R}(\Phi(e_i\cdot m_i^{proj})).
\end{equation}

Finally, we encode the observation of system states $s^{sys}_1,s^{sys}_2,s^{sys}_3$ for the third layer of our representation. 
Specifically, we use three non-overlapping rectangle masks $m_i^{sys}$ to encode scalar signals of system states to an image.
These masks, of size $N_{sys} \times 64$, span entire rows of a $64 \times 64$ image, where $N_{sys} \in \left(0, \frac{64}{3}\right]$ denotes the number of rows in the mask.
The normalized states are multiplied by the rectangular masks.
Finally, we multiply the resultant images by $255$ (the UInt8 range) and sum them to obtain the third layer of the DSA image, which is
\begin{equation}
    I_3^{DSA} = 255\sum_{i=1}^{3} s^{sys}_i \cdot m_i^{sys};
\end{equation}

After obtaining 3 layers, we stack them to create a DSA image as $I^{DSA} = [I_1^{DSA},I_2^{DSA}, I_3^{DSA}]\in\mathbb{R}^{64\times 64\times3}$, which is further used as the input of our visuomotor policy.

 \begin{figure*}[!tbp]
  \centering
  \includegraphics[width=1.0\hsize]{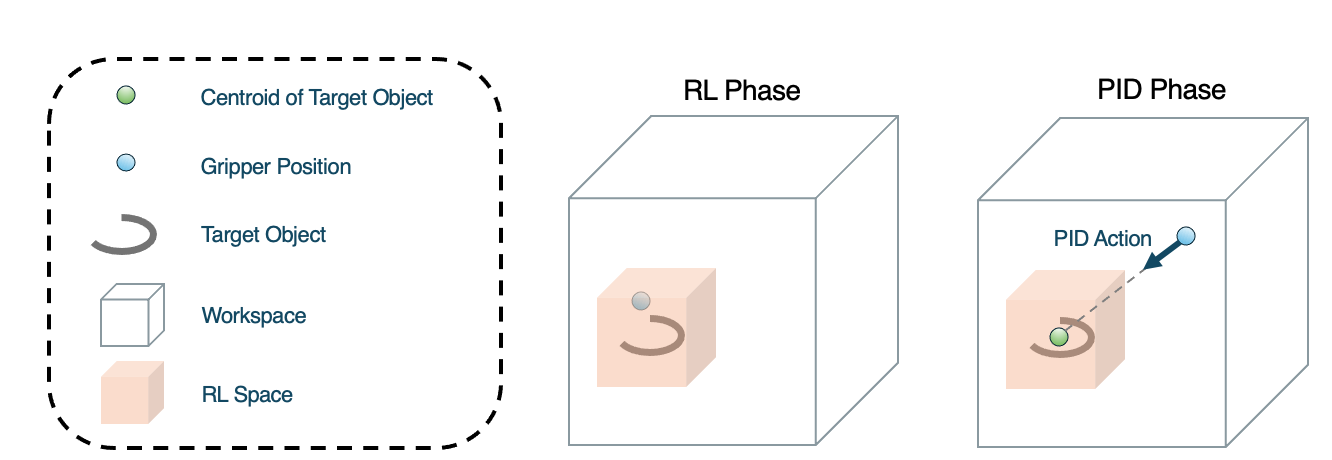}
  \caption{PID phase vs. RL phase. When the robot gripper is outside an RL space, centered at the centroid of the target's voxels, then the controller is at the PID phase. Otherwise, the controller is at the RL phase. For the PID phase, the action direction of the PID policy is always pointed to the centroid of the target's voxels.} 
  \vspace{-0.45cm}
\label{fig:hybrid_control}
\end{figure*}

 \begin{figure}[!tbp]
  \centering
  \includegraphics[width=1.0\hsize]{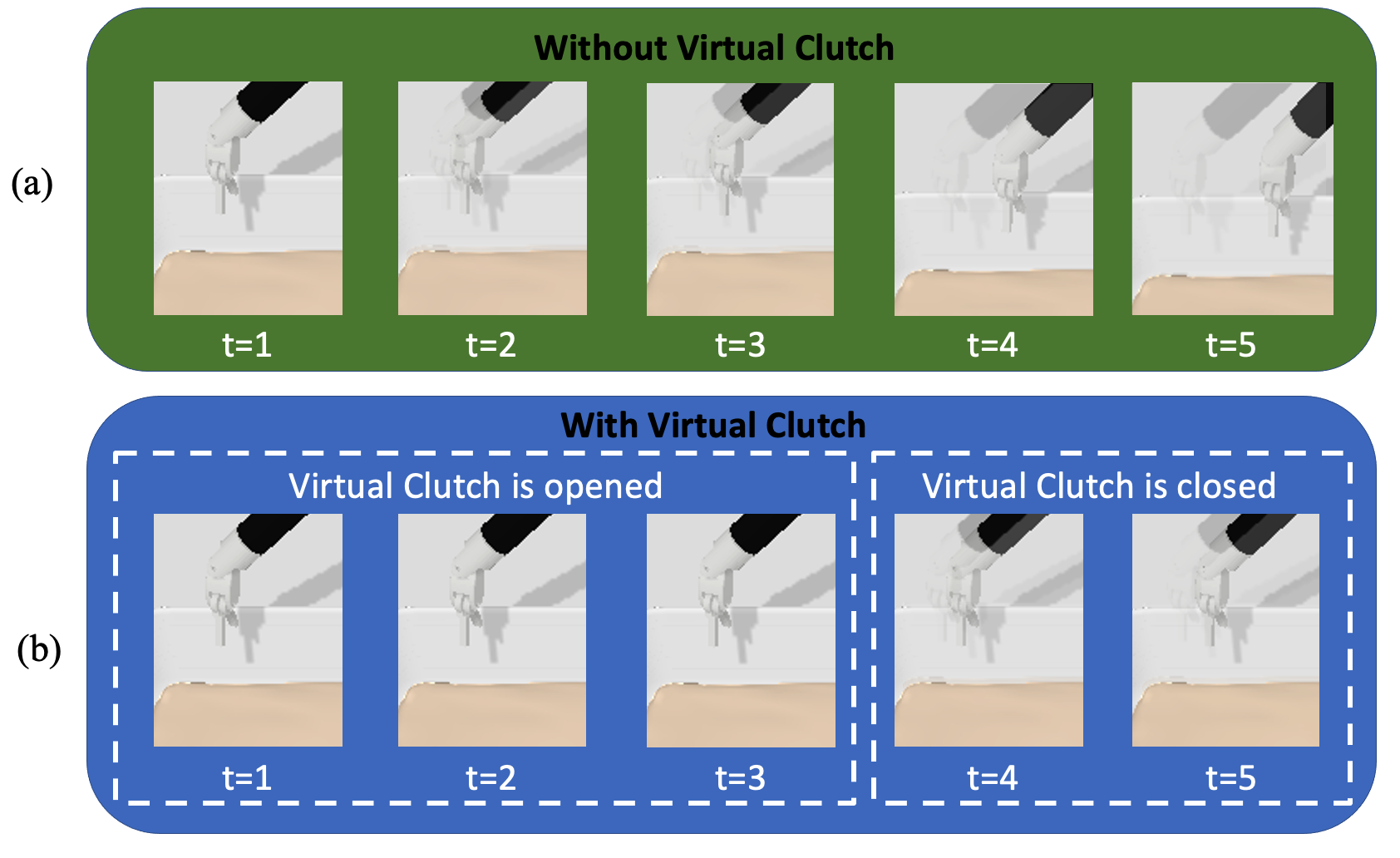}
  \caption{\color{black}Schematic illustration for virtual clutch (VC). (a) We show the first 5 timesteps of a rollout when a robot gripper is controlled to move to the right without applying VC. The transparent areas represent the starting positions of the gripper. (b) We show the corresponding 5 timesteps after applying VC ($H_{begin}=4$). The gripper remains stationary until the fourth timestep when it begins to move. \color{black}} 
  \vspace{-0.45cm}
\label{fig:VC}
\end{figure}

\subsection{Hybrid Control Architecture}
\label{sec:hybrid control}

Applying world-model-based visuomotor controllers to surgical grasping faces several challenges.
First, the success rewards within the POMDP's state space are extremely sparse due to the definition of discrete rewards, resulting in slow convergence of training and poor asymptotic performance. 
Second, in certain trivial cases, the trajectory generated by the learned visuomotor policy may deviate from the optimal trajectory due to the stochastic nature of the policy.
The sub-optimal performance in such trivial cases prevents practitioners from deploying RL policy to RAS, where safety and time efficiency are key priorities. 
Third, our recurrent RL policy performs poorly at the beginning of a rollout. 
This is primarily due to the fact that the world model of DreamerV2 starts from a blank state \cite{wu2023daydreamer, hafner2020mastering} at the beginning of each rollout, resulting in a large estimation error for the latent states of the world model.

To deal with the above challenges, we propose a hybrid control architecture that combines traditional PID control, our previously proposed control of \textit{virtual clutch} (VC) \cite{lin2024world}, and our RL policy. 
We start with introducing the classification of control phases, followed by elaborating on sub-policies for the corresponding phases.
Finally, we introduce a safe control design for handling platforms with height variance.

\subsubsection{Control Phase Classification}
\label{sec:control_phase}

We identify three phases for a trajectory: i) \textit{beginning phase}, ii) \textit{PID phase}, and iii) \textit{RL phase}.
For the beginning phase, our visuomotor controller do not observe enough temporal observations.
We identify transitions with the timesteps $t<H_{begin}$ in such a phase, where $H_{begin}\in[0,H]$ is a non-negative constant that determines the timestep at which the controller obtains adequate temporal observations. 
When the timestep $t \geq H_{begin}$, the transitions are classified as either the PID phase or the RL phase. 
We classify them according to the spatial distance between the gripper and the target object. 
In particular, we measure the distance between the centroids of the identified voxels for the gripper and the target grasping object, calculated in (\ref{equ:voxel_centroids}). 
Let $(c_x^{1},c_y^{1},c_z^{1})$ and $(c_x^{2},c_y^{2},c_z^{2})$ be the centroid coordinates of identified voxels for the gripper and the target grasping object, respectively.
We define a distance metric vector $l_{dis}$ as
\begin{equation}
\begin{aligned}
    l_{dis} = \begin{bmatrix}
c_x^{1} - c_x^{2}-L^{dis}_x\\
c_y^{1} - c_y^{2}-L^{dis}_y\\
c_z^{1} - c_z^{2}-L^{dis}_z
\end{bmatrix},
\label{equ:pid_condition}
\end{aligned}
\end{equation}
where $L^{dis}_x, L^{dis}_y, L^{dis}_z \in \mathbb{R}$ are offsets w.r.t $x$, $y$ and $z$ axes.
We define the transition as the RL phase when the absolute value of the elements in $L_{dis}$ is smaller than a positive threshold $C_{dis} \in \mathbb{R}^+$, which can be expressed as $|l_{dis}| < C_{dis}$.
Otherwise, when the absolute value of the elements is equal to or larger than the positive threshold, i.e., $|l_{dis}| \geq C_{dis}$, the transition is defined as the PID phase.
The RL space in Fig. \ref{fig:hybrid_control} visualizes the Cartesian space identified as the RL phase using our identification rules.

\subsubsection{Sub-policies}
\label{sec:control_subpolicy}
We have three sub-policies corresponding to three control phases in our hybrid control architecture. 

For the beginning phase, since the temporal observations are insufficient for our RL controller, we control the robot using a safe trajectory to prevent interaction with the environment, thereby avoiding catastrophic outcomes.
We employ an idle action that keeps the robot's joint positions unchanged.
The idle action is defined as $a = [0, 0, 0, 0, 1]^T$, ensuring that the gripper's pose remains fixed and its jaw stays open.
For the PID phase, we use a PID policy based on the centroid coordinates of identified voxels in (\ref{equ:voxel_centroids}).
The action of the PID policy, denoted as $a=[a_x^{pid},a_y^{pid},a_z^{pid},0,1]^T$, drives the gripper to move toward the target object while maintaining a fixed orientation and an open jaw, where $a_x^{pid},a_y^{pid},a_z^{pid}\in \mathbb{R}$ are three PID action variables to actuate the gripper in $x,y$ and $z$ axes, respectively.
The PID action variables are determined by the differences between the centroid coordinates, expressed as
\begin{equation}
\begin{aligned}
    a^{pid}_x &= f_{clip}(K_p\cdot(c_x^{1} - c_x^{2})) \\
    a^{pid}_y &= f_{clip}(K_p\cdot(c_y^{1} - c_y^{2})) \\
    a^{pid}_z &= f_{clip}(K_p\cdot(c_z^{1} - c_z^{2})),
\end{aligned}
\end{equation}
where $K_p$ is a positive constant representing the PID's proportional gain, and $f_{clip}$ is a clip function that saturates the input to the normalized action range of $[-1, 1]$. 
The actuated direction of our PID action is visualized in Fig. \ref{fig:hybrid_control}.

For the RL phase, the output of our learned visuomotor policy is directly used for the control command. 
We use two hyper-parameters $\alpha^{xyz},\alpha^\theta\in(0,1]$ to shrink the magnitudes of the actuated robot command for discrete action.
In particular,  the maximum translational and rotational magnitudes are shrunk to $\alpha^{xyz}\Delta^{xyz}$ and $\alpha^\theta\Delta^{\theta}$,  respectively.
Thus, the actuated action command for our RL policy can be represented as $a=[\alpha^{xyz},\alpha^{xyz},\alpha^{xyz},\alpha^\theta,1]^T \odot  \hat{a}$.

\subsubsection{Safe Control for Handling Platforms with Height Variance}
\label{sec: Safe Control for Handling Platforms with Height Variance}
Although our controller with the hybrid-control architecture can effectively actuate the gripper to grasp a target object, there is no guarantee that the gripper will avoid hitting the platform beneath the target. This presents a significant risk in real surgical settings, as any contact with the platform could potentially damage critical and fragile tissues of surgical platforms.

To mitigate this risk, it is essential to constrain the gripper's movement, ensuring that it does not damage the platform while still achieving the grasping goal. Our observation is that in typical RAS settings, the height (z-coordinate) of the target object is generally higher than that of the platform.

Building on this observation, we propose a simple yet effective method to prevent the gripper from damaging the platform. Specifically, we use the z-centroid coordinates of the identified voxels for both the gripper, $c_z^{1}$, and the target object, $c_z^{2}$, as our observation.
If the difference in z-coordinates between the gripper and the target object $c_z^{1}- c_z^{2}$ becomes smaller than a predefined safe threshold, $Z_{safe}$, after the gripper is actuated by our hybrid-architecture controller, an additional corrective move is triggered. The gripper will move upward---along the positive direction of the z-axis---by $30\text{mm}$.
This method ensures that the gripper maintains a certain height relative to the target object. Since the platform is typically positioned below the target, the minimum height of the gripper relative to the platform is greater than or equal to $Z_{safe}$, thus preventing the gripper from applying excessive force to the platform's tissue. We apply this method in our real-robot experiments.

\subsection{Training Scene for Sim-to-Real Transfer}
Training a visuomotor policy in simulation and transferring it to a real surgical robot alleviates the need for data collection on real robots, thus reducing operational costs and human labor.
In the following, we introduce our simulation training environment.
We also incorporate a tilted gripper pose to enhance visibility, along with domain randomization to improve the sim-to-real transfer.

\begin{figure}[!tbp]
  \centering
  \includegraphics[width=\hsize]{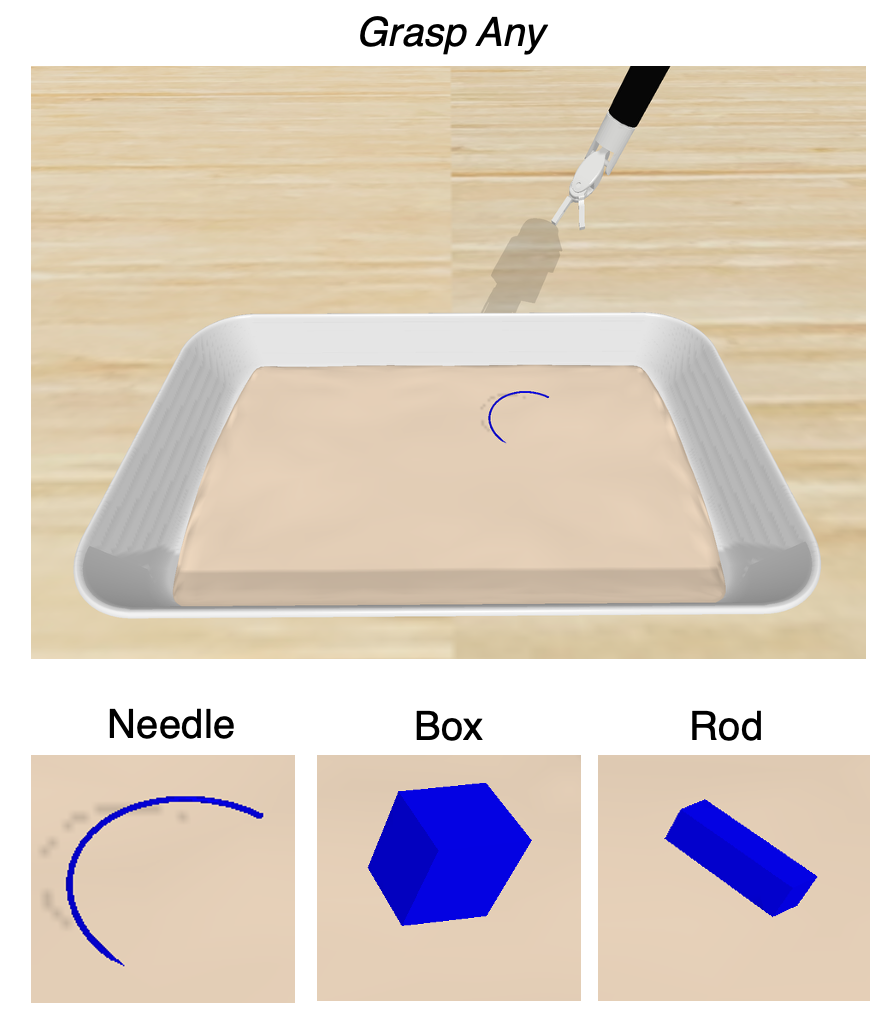}
  \caption{Our training task in simulation, Grasp Any. The top figure illustrates the setup for our grasping task, while the bottom figure displays the target object used in the simulation.}
  \vspace{-0.45cm}
\label{fig:exp_setup}
\end{figure}

\subsubsection{Grasp Any}
Our visuomotor controller is trained in simulation scenes developed based on an SOTA simulation platform for surgical RL learning, SurRoL \cite{xu2021surrol}. 
Our training task (see Fig. \ref{fig:exp_setup}), \textit{Grasp Any}, is similar to the \textit{NeedlePick} task in SurRoL.
The robotic arm is a 6-DOF manipulator, \textit{patient side manipulator} (PSM).
The stereo camera remains fixed during each rollout, with its frame’s $z$-axis tilted $20^\circ$ from the vertical relative to the ground.
The intrinsic and extrinsic parameters of the stereo camera are directly provided in the simulation. 
A $20$mm needle is used as the target object. 
The depth images and segmentation masks from the camera are also directly rendered in simulation, thereby we do not need to apply depth estimation (in Section \ref{sec:depth estimation}) and segmentation (in Section \ref{sec:segmetation}) in simulation. 

Compared to the original NeedlePick task, we enhance the difficulty of our learning task for our visuomtor learning.
First, the initial poses of both the gripper and the target object are randomized within the workspace.
Notably, the initial height of the target object is randomized instead of a fixed height in the NeedlePick task. 
Second, the task is terminated after the gripper's jaw is closed for the first time, as defined in Sec. \ref{sec: Termination and Reward Design}.
This approach allows the robot agent to accelerate task completion, given the limited number of grasp trials---one in our case.
Third, the robot's joint measurement or an object's Cartesian pose is not observable according to our assumption. 
The agent has to reason the object's spatial information purely from visual observation.
Finally, to improve the generality of visuomotor policy, we add two more novel objects besides the needle: a block and a rod (See fig. \ref{fig:exp_setup}). 
The occurrence probabilities of the needle, the block, and the rod are $0.5$, $0.25$, and $0.25$, respectively.

 \begin{figure}[!tbp]
  \centering
  \includegraphics[width=0.8\hsize]{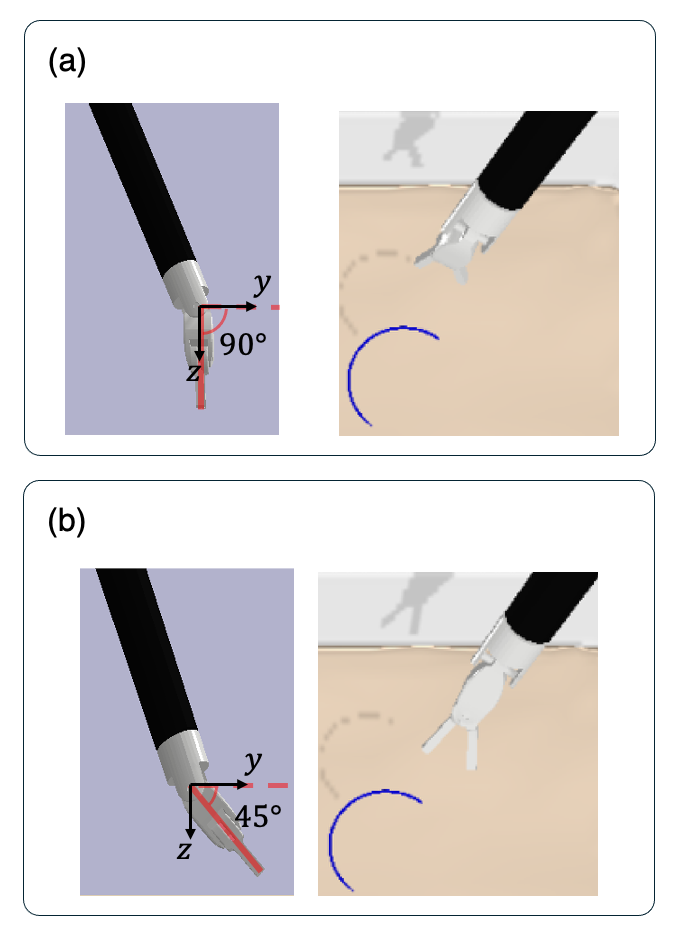}
  \caption{Conventional gripper orientation vs. our gripper orientation. a) The left image shows the conventional gripper orientation, where its $z$-axis is perpendicular to the ground plane. The right image displays the corresponding RGB image from the left stereo camera.
b) The left image illustrates the orientation of our gripper, with its $z$-axis tilted 45 degrees relative to the ground. The right image shows the corresponding RGB image, with increased visibility compared to the conventional orientation.} 
  \vspace{-0.45cm}
\label{fig:gripper_orientation_comparison}
\end{figure}
\subsubsection{Enhancing Gripper visibility}
To increase the visibility of the gripper,  we modify the orientation in $x$ axis of our gripper compared to the original needle-picking task. 
A conventional orientation of the gripper ensures that the gripper's $z$ axis is always perpendicular to the ground.
However, this configuration would result in a lack of visibility of the gripper's tip, as the canonical stereo cameras in RAS are typically set in a near top-down view.
Fig. \ref{fig:gripper_orientation_comparison}a demonstrates the conventional gripper orientation and its corresponding image view.
To address this issue, we apply a tilted orientation for our gripper.
In particular, we rotate the gripper by 45 degrees around the $x$-axis of its own reference frame.
This ensures that the gripper's $z$-axis remains at a 45-degree angle to the ground.
Our tilted orientation enhances the visibility of the gripper's tip (see Fig. \ref{fig:gripper_orientation_comparison}b), thereby improving the overall performance of our visuomotor policy.
We apply our tilted orientation in both simulation and real robots.

\begin{table}[!tbp]
\scriptsize
\renewcommand{\arraystretch}{1.3}
\caption{Value/Range for Domain Randomization}
\centering
\begin{tabular}{c c c }
   \Xhline{4\arrayrulewidth}
   \bf{Type} & \bf{Name} & \bf{Value/Range}  \\ \hline
   \multirow{4}{*}{Random Cam Pose} & Roll Angle Noise & $[-3^{o}, 3^{o}]$ \\
   & Pitch Angle Noise& $[-3^{o}, 3^{o}]$ \\
   & Yaw Angle Noise& $[-1^{o}, 1^{o}]$ \\
   & Distance Noise & $[-10mm, 10mm]$ \\
   \hline
    \multirow{1}{*}{Object Scaling} 
    &  Object Scaling & $[0.75, 1.25]$ \\
    \hline
    \multirow{1}{*}{Action Noise} & Noise Scaling & $[-0.01, 0.01]$ \\
    \hline
    \multirow{4}{*}{Image Noise}
     & Uniform Noise Range & $0.005$ \\
     & Gaussian Blur Kernel Size& $3$ \\
     & Gaussian Blur Sigma& $0.3$ \\
     & Image Cutout Amount& $[0.0,0.2]$ \\

   \hline
\Xhline{4\arrayrulewidth}
\end{tabular}
\label{table:domain_random}
\end{table}
\begin{figure*}[!tbp]
  \centering
  \includegraphics[width=1\hsize]{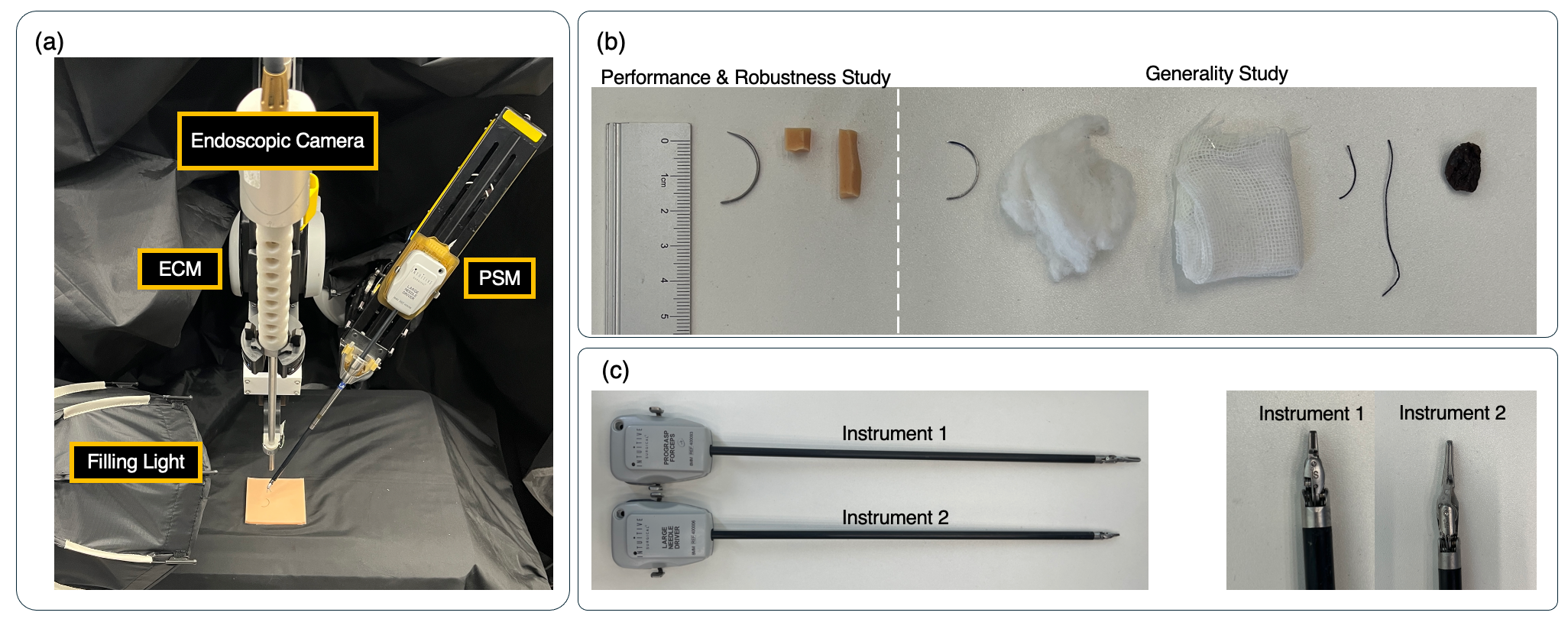}
  \caption{Robot setup, target objects, and grippers in real robot experiments. (a) The real-world robotic setup used in our experiments. (b) The set of target objects used to evaluate performance, generalization, and robustness. (c) The grippers used in our robotic system, each actuated by specialized instruments.}
  \vspace{-0.45cm}
  \label{fig:surgical_objects}
\end{figure*}
\subsubsection{Domain Randomization}
To enhance both the performance and robustness of our visuomotor policy for sim-to-real transfer, we deploy domain randomization in our simulation scene.
In particular, we apply four types of domain randomization:  
\begin{itemize}
\item \textit{Random Cam Pose}: We inject noise into the camera pose, compelling our learned visuomotor controller to adapt to small errors in the extrinsic parameters of the camera calibration, similar to the approach in \cite{seo2023multi}.
Specifically, we add noise to the 3D orientation of the camera.
We also introduce noise to the distance between the camera and the center of the workspace.
These noises remain fixed during a rollout.
Note that the noise is unobservable by our controller.

\item \textit{Object Scaling}: We scale the object size to improve the controller's generality in terms of the variance of the target object's size. 
The size of the target grasping object is randomly scaled within a range. 

\item \textit{Action Noise}: We add noise to the robot command to simulate the actuation error between the measured and desired gripper positions under end-effector control in real robots, primarily caused by hysteresis and slacks in the robot system.
For each timestep, noises with a uniform distribution are added to the Cartesian position of the actuated target pose after applying the controller's action. 
The noise magnitude is proportional to $\Delta^{xyz}$, controlled by a scaling factor.

\item \textit{Image Noise}: Simulated image noises are added to the observed images in simulation, as \cite{horvath2022object} suggests. 
In particular, to simulate the depth noises caused by our depth estimation in real robots, we add noise with a uniform distribution to the pixel value in depth images in simulation. 
In addition, to simulate the image blur caused by fogging during surgery, Gaussian blur is applied to the depth image.
The blur is controlled by the size of the Gaussian kernel and its variance.
Finally, we simulate both occlusions and segmentation errors in real robots using cutouts.
In particular, we apply cutouts to the segmentation masks. 
For each mask, square and circle cutouts with random coordinates and sizes are applied to the object's mask.
We control the cutouts by the cutout amount, which is the ratio of cutout pixels in the mask area.
An example of image noise is visualized in Fig. \ref{fig:gasv2_overview}a.

\end{itemize}
\color{black}Detailed parameters of our domain randomization are shown in Table \ref{table:domain_random}.\color{black}

\section{Experiment Setup}
\subsection{Evaluation Scene}
We evaluate our method on both simulation and real robots. 
For simulation, we use the same experiment setup in our training scene, except for using different random seeds. 
For real robots, we deploy our controller to a surgical system (see Fig. \ref{fig:surgical_objects}a), including a PSM for the grasping robotic arm and an actuated camera arm, \textit{endoscopic camera manipulator} (ECM), mounted with a stereo endoscopic camera.
The robotic arms are controlled by an open-source control system, da Vinci Research Kit (dVRK) \cite{kazanzides2014open}.
Stereo images are captured by the endoscopic stereo camera with 1080p resolution.
The images are center cropped to the size of $600\times600$ for the controller's observation. 
A filling light is used to control the illumination condition of environment.
On the real robot, the camera pose is chosen to lie within the domain-randomized range used in simulation.
For each rollout, we randomize the initial pose of a target object within the workspace manually, while the gripper is actuated to a randomized initial pose.
The task will be terminated when the timestep reaches the maximum limit or the gripper is closed.
When the gripper closes, the task is considered complete, and the gripper is then lifted by $30$ mm.
If the gripper is holding an object after being lifted, the FSM state is labeled as the successful termination.
Otherwise, it is labeled as the failed termination.
Domain randomization is not applied during real-world robot deployment.

\subsection{Baselines and Ablations}
SOTA pixel-level RL methods, \texttt{PPO} \cite{schulman2017proximal} and \texttt{DreamerV2} \cite{hafner2019dream} are compared in our experiments. 
In these baselines, visual signals (i.e., RGB-D observed images, object masks) and two system signals (i.e., gripper states and task-level states) are directly fed as observations of the visuomotor controllers.
To represent depth images and masks in our baseline comparisons, we use a visual encoding method called Raw Visual Representation (RawVR).
This consists of three stacked image layers: (i) a depth layer, (ii) a mask layer, and (iii) a system state layer. 
The depth layer is created by summing all depth segments. 
The mask layer combines all masks sharing the same encoded values defined by our DSA. 
The system state layer sums encoded images of two system signals using the same encoding method in our DSA.
For consistency, the final stacked image is downsampled to a resolution of $64\times 64 \times 3$.
For PPO, the resultant image is encoded to a feature vector via a pre-trained universal visual encoder, R3M \cite{nair2022r3m}; a multilayer perception (MLP) policy leverages the vectorized encoding as input and is learned by the standard PPO framework \cite{schulman2017proximal}.
For DreamerV2, it is using the same learning world-model-based framework, without components of our visual perception and hybrid-control architecture\cite{hafner2020mastering}.
We also include a comparison with our previous work, \texttt{GAS} \cite{lin2024world}, which represents the SOTA visuomotor baseline for general grasping tasks. 
GAS shares the same visuomotor learning setup as GAV2, with two key differences: it uses a distinct visual representation based on DSA and does not incorporate the hybrid control architecture.

To evaluate the effectiveness of individual components of our approach, we compare with ablation baselines as follows:
\begin{itemize}
   \item \texttt{GASv2-RawVR}: 
   \color{black}
   We replace our visual representation with RawVR to evaluate the effective of our visual representation. \color{black}
   \item \texttt{GAS-NoClutch}: We remove the effect of VC to evaluate its effectiveness, which can be achieved by setting $H_{begin}$ to $0$.
   \item \texttt{GAS-NoDR}: We remove the effect of domain randomization during training to evaluate its effectiveness in sim-to-real transfer.
   \item \texttt{GAS-NoPID}: We remove the effect of PID subpolicy in our hybrid control architecture. In this baseline, the PID phase is removed in the hybrid control architecture.
\end{itemize}

\subsection{Training and Evaluation}

For performance evaluation, we adopt two metrics: (i) the success rate, a widely used benchmark in surgical grasping \cite{xu2021surrol,chiu2021bimanual,wilcox2022learning}, and (ii) the grasping score ($s_{g}$), a novel metric designed to quantify the efficiency of the controller in achieving a successful grasp.
 We define the score $s_g\in[0,1]$ as 
\begin{equation}
\label{equ:score}
     s_{g} = \begin{cases} \frac{(H - H_{max})}{H_{max}} &  \text{if the task succeed} \\
    0 & \text{otherwise}
    \end{cases};
\end{equation}
where $H$ is the terminated timestep of a rollout.

In the training process, controllers are trained with $1.1$ million timesteps; 
For each baseline, $20$ evaluation rollouts are carried out every $40\text{K}$ training timestep; 
we pre-fill $100$K timesteps of rollouts before training, including $50$K timesteps using a scripted demonstration policy and $50$K timesteps with a random policy, as we mentioned in Section \ref{sec:Network Architecture and Training}. 
We train on a computer with 4 GPUs (Nvidia RTX 3090) and a 16-core CPU. 
Each baseline costs $6$ days on the computer for training.
In the real robot, the evaluated controllers run at around 1 Hz, where the computer configuration is a desktop with a 24G-RAM GPU, Nvidia TITAN RTX, and an 8-core CPU. 
The detailed hyperparameters used for training are provided in Table \ref{table:hyperparam}.

 \begin{figure}[!tbp]
  \centering
  \includegraphics[width=\hsize]{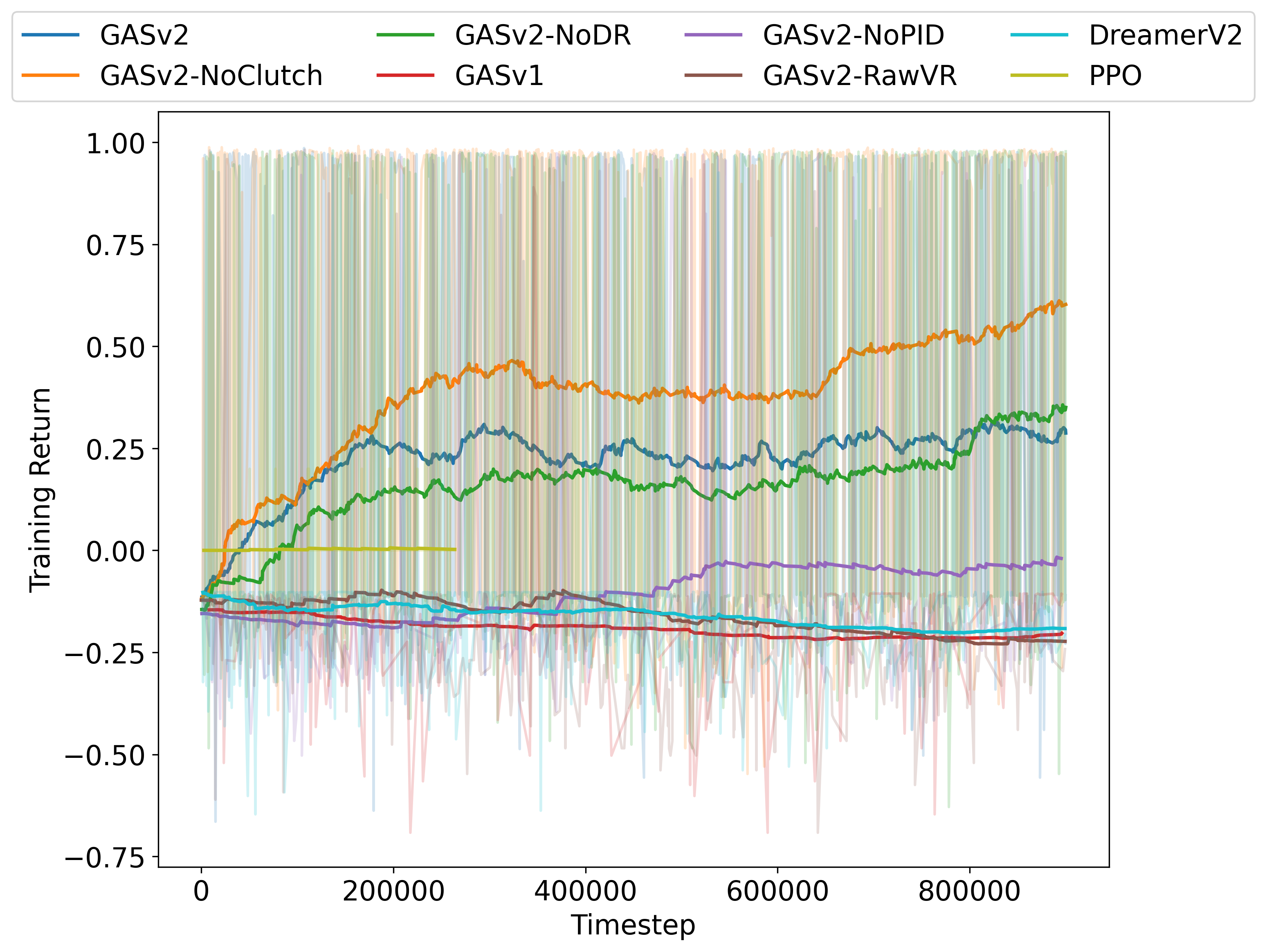}
  \caption{Returns of training visuomotor policies in simulation. Curves are smoothed by Exponential Moving Window with a $0.99$ smoothing factor.}
  \vspace{-0.45cm}
  \label{fig:training_curve}
\end{figure}

\begin{table*}[htbp]
\centering
\begin{threeparttable}
\renewcommand{\arraystretch}{1.3}
\fontsize{8pt}{8pt}\selectfont
\caption{Success rate (\%) in simulation experiments over 5 seeds}
\label{table:simulation_success_rate}
\begin{tabular}{|l|l|c|c|c|c|c|c|c|}
\hline
\multirow{2}{*}{\textbf{Task Category}} & \multirow{2}{*}{\textbf{Specific Task}} & \multicolumn{7}{c|}{\textbf{Baselines}} \\ \cline{3-9} 
 &  & \textbf{GASv2} & \textbf{GASv2-RawVR} & \textbf{GASv2-NoClutch} & \textbf{GASv2-NoPID} & \textbf{DreamerV2} & \textbf{PPO} & \textbf{GASv1} \\ \hline
\multirow{1}{*}{\textbf{Performance}} & Performance Study & 53.8 (5.2) & 3.0 (0.6) &  \bf{62.0 (6.6)} & 41.4 (4.7) & 0.8 (0.4) & 0.0 (0.0) & 0.2 (0.4)\\ \hline
\multirow{3}{*}{\textbf{Generalization}} & OOD Large & \bf{54.0 (1.7)} & 5.6 (3.0) & 52.4 (7.6) & 39.8 (5.6) & 0.8 (0.4) & 0.2 (0.4) & 0.2 (0.4) \\ \cline{2-9}
 & OOD Small & \bf{49.8 (4.1)} & 2.6 (1.0) & 48.2 (4.5) & 19.8 (3.7) & 0.4 (0.5) & 0.0 (0.0) & 0.0 (0.0) \\ \cline{2-9}
 & OOD Shape & \bf{52.0 (4.9)} & 1.2 (0.7) & 47.4 (5.6) & 26.0 (2.8) & 0.2 (0.4) & 0.0 (0.0) & 0.0 (0.0) \\ \hline
\multirow{2}{*}{\textbf{Robustness}} & Moving Camera & 56.8 (2.2) & 5.2 (1.7) & \bf{62.0 (3.0)} & 42.8 (1.9) & 1.0 (0.6) & 0.0 (0.0) & 0.0 (0.0) \\ \cline{2-9}
 & Re-grasp & 58.0 (2.4) & 2.8 (0.7) & \bf{62.0 (4.9)} & 36.6 (3.6) & 1.0 (1.5) & 0.0 (0.0) & 0.0 (0.0) \\ \hline \hline
 \multicolumn{2}{|c|}{\textbf{Average}}  & 54.1 (4.6) & 3.4 (2.2) & \bf{55.7 (8.6)} & 34.4 (9.4) & 0.7 (0.8) & 0.0 (0.2) & 0.1 (0.2) \\ \hline
\end{tabular}
\begin{tablenotes}
\footnotesize
\item \textit{Note:} The table shows the mean and standard deviation of success rates (in \%) over $5$ random seeds, each evaluated using $100$ rollouts. Results are presented as mean (standard deviation). Best-performing values among baselines are highlighted in bold.
\end{tablenotes}
\end{threeparttable}
\end{table*}

\begin{table*}[htbp]
\centering
\begin{threeparttable}
\renewcommand{\arraystretch}{1.3}
\fontsize{8pt}{8pt}\selectfont
\caption{Grasping score (\%) in simulating experiments over 5 seeds}
\label{table:simulation_grasping_score}
\begin{tabular}{|l|l|c|c|c|c|c|c|c|}
\hline
\multirow{2}{*}{\textbf{Task Category}} & \multirow{2}{*}{\textbf{Specific Task}} & \multicolumn{7}{c|}{\textbf{Baselines}} \\ \cline{3-9} 
 &  & \textbf{GASv2} & \textbf{GASv2-RawVR} & \textbf{GASv2-NoClutch} & \textbf{GASv2-NoPID} & \textbf{DreamerV2} & \textbf{PPO} & \textbf{GASv1} \\ \hline
\multirow{1}{*}{\textbf{Performance}} & Performance Study & 30.5 (29.2) & 1.1 (6.9) & \bf{40.4 (32.8)} & 16.8 (23.6) & 0.8 (0.4) & 0.0 (0.0) & 0.1 (2.2)\\ \hline
\multirow{3}{*}{\textbf{Generalization}} & OOD Large & 30.9 (29.4) & 1.9 (9.0) & \bf{34.2 (33.9)} & 17.1 (23.9) & 0.4 (4.0) & 0.2 (3.5) & 0.1 (1.7) \\ \cline{2-9}
 & OOD Small & 28.5 (29.4) & 1.3 (7.8) & \bf{31.5 (33.7)}& 8.5 (18.9) & 0.2 (2.1) & 0.0 (0.0) & 0.0 (0.0) \\ \cline{2-9}
 & OOD Shape & \bf{29.7 (29.4)} & 0.3 (3.2) & 29.1 (31.8) & 10.5 (19.9) & 0.1 (1.8) & 0.0 (0.0) & 0.0 (0.0) \\ \hline
\multirow{2}{*}{\textbf{Robustness}} & Moving Camera & 32.5 (29.4) & 1.5 (7.6) & \bf{40.5 (32.8)} & 18.5 (24.5) & 0.4 (4.4) & 0.0 (0.0) & 0.0 (0.0) \\ \cline{2-9}
 & Re-grasp & 33.0 (29.1) &1.2 (7.4) & \bf{40.8 (33.1)} & 15.5 (23.1) & 0.4 (4.1) & 0.0 (0.0) & 0.0 (0.0) \\ \hline \hline
 \multicolumn{2}{|c|}{\textbf{Average}}  & 30.9 (29.4) & 1.2 (7.2) & \bf{36.1 (33.4)} & 14.5 (22.7) & 0.3 (3.7) & 0.0 (1.6) & 0.0 (1.1) \\ \hline
\end{tabular}
\end{threeparttable}
\begin{tablenotes}
\footnotesize
\item \textit{Note:} The table shows the mean and standard deviation of the grasping score (in \%) over $5$ random seeds, each evaluated using $100$ rollouts. Results are presented as mean (standard deviation). Best-performing values among baselines are highlighted in bold.
\end{tablenotes}
\end{table*}
\section{Simulation Results}
\subsection{Performance}
We evaluate the performance of SOTA baselines---GAS, PPO, and DreamerV2---against our proposed method.
Figure \ref{fig:training_curve} illustrates the training return curves in the simulated environment.
As training progresses, the return increases steadily with time steps and converges to an average of approximately $0.65$ by the end of training.
In contrast, GAS, PPO, and DreamerV2 show no improvement during training and converge to an average return of approximately $-0.1$.
This outcome may be due to the high complexity of the task and the limited number of training steps.

Tables \ref{table:simulation_success_rate} and \ref{table:simulation_grasping_score} show the success rate and grasping score in the evaluation of the simulation, respectively. 
The results are evaluated using $5$ random seeds, with each seed evaluated over $100$ evaluation rollouts.
Our approach achieve strong performance on simulation evaluation, with an average of success rate of $53.8\%$ and a grasping score of $30.5\%$.
In contrast, the controllers of DreamerV2, PPO GASv1 perform poorly in the evaluation experiments.
Our approach achieves strong performance in simulation, with an average success rate of $53.8\%$ and a grasping score of $30.5\%$.
In contrast, the controllers based on DreamerV2, PPO, and GASv1 perform poorly in the evaluation experiments, each achieving both the success rate and grasping score of less than $1\%$.
In summary, our approach significantly outperforms the SOTA baselines in both training and evaluation within the simulated environment.

\subsection{Generality}
The simulation experiments are designed to evaluate the generalization capability of the controllers to OOD object size and shape.
We define two simulation scenarios---OOD Large and OOD Small---to assess generalization to the OOD sizes of target grasping objects.
In these settings, the target object is uniformly scaled to size ranges not seen during training: $[1.5,2.0]$ for the OOD-Large scenario and $[0.5,0.75]$ for the OOD-Small scenario.
Notably, the types of target objects and their occurrence probabilities remain consistent with those in the training environment.
In addition, we define another simulation scenario---OOD Type---to evaluate the generality under unseen object shapes.
A sphere object with a random size, which is not seen during training, is used as the target object for evaluation.
Notably, other grasping objects do not occur in this evaluation.

As shown in Tables \ref{table:simulation_success_rate} and \ref{table:simulation_grasping_score}, our controller demonstrates strong generalization to OOD sizes and shapes of target grasping objects.
Across all OOD scenarios, it achieves a success rate exceeding $49.8\%$ and a grasping score above $28.5\%$.
Compared to the standard performance evaluation, the controller experiences only a slight degradation, with a maximum decrease of $4\%$ in success rate and $2\%$ in grasping score.
In the OOD Large scenario, our controller achieves a slight performance improvement—an increase of $0.2\%$ in success rate and $0.4\%$ in grasping score.
This increase is primarily due to the fact that grasping larger target objects poses fewer challenges for the controller in terms of control actuation and perceptual processing.
The SOTA baselines, DreamerV2 and GASv1, exhibit either slight performance degradation or no improvement across all OOD scenarios.
In contrast, PPO demonstrates a modest performance gain in the OOD Large setting, likely due to the reduced task difficulty presented by larger target objects.
All SOTA baselines achieve less than $0.8\%$ in both success rate and grasping score, indicating limited generalization to OOD variations in target object size and shape.

\subsection{Robustness}
We conduct two types of robustness studies---i) dynamic error of camera pose and ii) regrasping---in simulation to assess the robustness of the controllers.
The first study evaluates the controllers’ ability to adapt to dynamic errors in calibrated camera poses during the operation of RAS.
Such errors normally occur when the camera arm, ECM, moves during surgery.
While our training environment assumes fixed camera noise within a rollout, this robustness study introduces dynamic noise to simulate the variability encountered in real robotic systems.
The dynamic error is clipped to the range of Random Cam Pose in the domain randomization of our training scene.
The second study investigates the controllers’ ability to perform regrasping.
This capability is essential for a robust control system, as it allows recovery from failure.
While the task is terminated after the first grasp attempt in the training environment, we modify the termination conditions in the robustness study. Specifically, the task ends when: (i) the robot performs a second grasp attempt, (ii) the object is successfully grasped on the first attempt, or (iii) the maximum number of timesteps---identical to that of the training environment---is reached.
 
Our approach demonstrates strong robustness to camera pose disturbances and grasping failures, as shown in Tables \ref{table:simulation_success_rate} and \ref{table:simulation_grasping_score}.
In the study of dynamic camera pose errors, our controller demonstrates improved performance, with a $3\%$ increase in success rate and a $2\%$ increase in grasping score compared to the standard performance evaluation.
This performance gain under the dynamic pose disturbances may be attributed to the domain randomization of camera poses deployed during training.
In the regrasping study, our controller also demonstrates improved performance, with an increase of approximately $5\%$ in success rate and $3\%$ in grasping score.
This performance gain indicates that our controller is capable of recovering from failed grasp attempts, highlighting its robustness in handling failure cases.
For SOTA baselines, PPO and GASv1 both experience a decrease in success rate and grasping score, indicating poor robustness when faced with camera pose disturbances and grasping failures.
DreamerV2 shows a slight increase of $0.2\%$ in success rate but a decrease of $0.4\%$ in grasping score across all robustness studies.
This suggests that DreamerV2 is more robust than PPO and GASv1 in terms of success rate.
However, it suffers from longer completion times when operating under disturbances.

\subsection{Ablation}
\label{sec:exp_ablation}
Ablation baselines---GASv2-RawVR, GASv2-NoClutch and  GASv2-NoPID---are investigated in simulation scenarios of training and evaluation to evaluate effectiveness of our visual representation and two subpolicies (the VC and PID subpolicies) in our hybrid control architecture.
During training, the returns for both GASv2-NoPID and GASv2-NoClutch progressively improve with the number of training steps. 
In contrast, GASv2-RawVR exhibits no noticeable improvement in training return throughout the training process (see Fig. \ref{fig:training_curve}).
In the simulation evaluation, GASv2-NoPID and GASv2-RawVR exhibit inferior performance across all evaluation scenarios compared to our controller, as Tables \ref{table:simulation_success_rate} and \ref{table:simulation_grasping_score} show.
GASv2-NoPID exhibits a slight performance degradation, with a $19.7\%$ drop in average success rate and a $15.5\%$ decrease in average grasping score. 
In contrast, GASv2-RawVR shows a substantial decline, with a $50.7\%$ reduction in average success rate and a $29.7\%$ drop in grasping score.
The performance declines observed in GASv2-NoPID and GASv2-RawVR highlight the effectiveness of the PID subpolicy in our hybrid control architecture and the importance of our visual representation, respectively.
GASv2-NoClutch achieves performance comparable to our controller, with a slight improvement of $1.6\%$ in success rate and $5.2\%$ in grasping score.
Notably, our approach consistently achieves a higher success rate than GASv2-NoClutch across all generalization studies. 
This suggests that GASv2-NoClutch exhibits reduced robustness when faced with OOD variations in the size and shape of target objects.
This limitation may stem from the controller in GASv2-NoClutch being overfitted to the specific object characteristics present in the training environment.
The observed decline in GASv2-NoClutch’s performance in generalization studies highlights the effectiveness of VC in our hybrid control architecture.

\section{Real-Robot Results}

\begin{figure}[!tbp]
  \centering
  \includegraphics[width=\hsize]{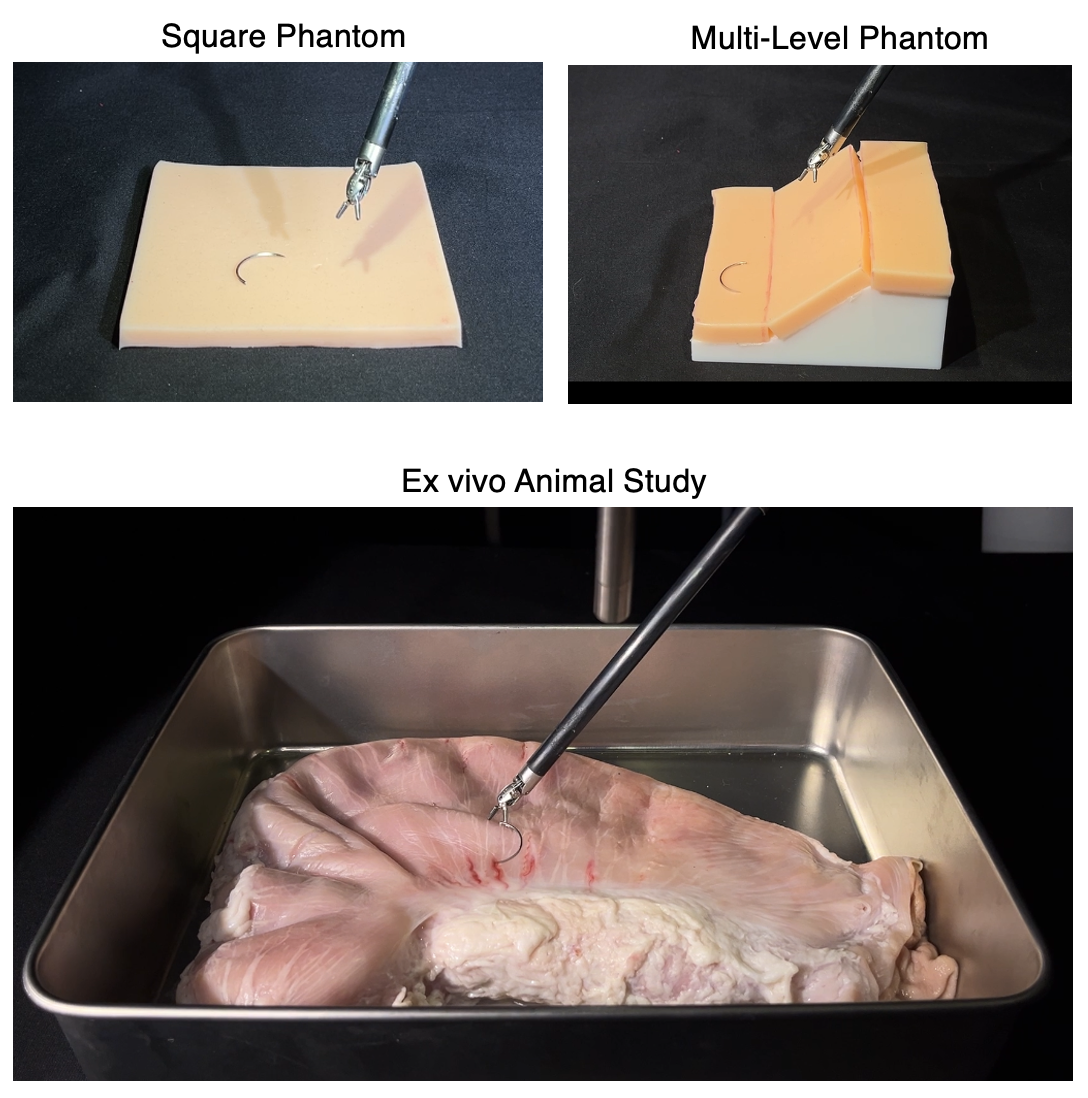}
  \caption{Platforms in our real-robot studies. This figure shows three platforms: a square phantom (top left) and a multi-level phantom (top right) for phantom-based studies, along with a porcine stomach (bottom) for our ex vivo animal study.}
  \vspace{-0.45cm}
\label{fig:real_robot_platforms}
\end{figure}

We conduct extensive experiments to evaluate the performance, generality, and robustness of the controllers on real robotic systems.
Grasping tasks are carried out on two types of phantom-based platforms---a square phantom and a multi-level phantom---as well as on an ex vivo animal platform using a porcine stomach, as shown in Fig. \ref{fig:real_robot_platforms}.
In the following, we elaborate on the experiment setup and results for these three types of platforms.

\begin{table*}[!tbp]
\centering
\scriptsize
\renewcommand{\arraystretch}{1.8}
\fontsize{8pt}{8pt}\selectfont

\begin{threeparttable}
    \caption{Real-Robot Experiments on Square Phantom for GASv2}
    \label{table:dvrk_square_phantom}

    \begin{tabular}{|c | c c c | c | c| c|}
        \Xhline{4\arrayrulewidth}  
        \bf{Study}  &\bf{Type} & \bf{Description} & \bf{Episode} & \bf{SR (\%)} & \bf{Type Weighted SR (\%)} & \bf{Study Weighted SR (\%)} \\
        \Xhline{2\arrayrulewidth}

        \multirow{3}{*}{Performance} 
            & Needle & 20mm Needle & 20 & 70 & 70 & \multirow{3}{*}{75} \\
        \cline{2-6}
            & Block & Phantom Block & 10 & 80 & 80 & \\
        \cline{2-6}
            & Rod & Phantom Rod & 10 & 80 & 80 & \\
        \Xhline{2\arrayrulewidth}

        \multirow{8}{*}{Generality} 
            & Needle &14mm Needle & 20 & 75 & 75 & \multirow{9}{*}{77.5} \\
        \cline{2-6}
            & \multirow{2}{*}{Thread} & Long Thread & 10 & 60 & \multirow{2}{*}{65} & \\
            & & Short Thread & 10 & 70 & & \\
        \cline{2-6}
            & Gauze & - & 20 & 95 & 95 & \\
        \cline{2-6}
            & Sponge & - & 20 & 100 & 100 & \\
        \cline{2-6}
            & Debris & Raisin & 20 & 90 & 90 & \\
        \cline{2-6}
            & OOD Gripper & Forceps Gripper & 20 & 65 & 65 & \\
        \cline{2-6}
            & \multirow{2}{*}{OOD Camera Pose} & Tilt Angle $0^{\circ}$ & 20 & 70 & \multirow{2}{*}{65} & \\
            & & Tilt Angle $40^{\circ}$ & 20 & 60 & & \\
        \Xhline{2\arrayrulewidth}

        \multirow{6}{*}{Robustness} 
            & \multirow{2}{*}{Background} & RAS Background 1 & 10 & 60 & \multirow{2}{*}{65} & \multirow{6}{*}{61} \\
            & & RAS Background 2 & 10 & 70 & & \\
        \cline{2-6}
            & Target Noise & Back\&Forth Movement & 20 & 55 & 55 & \\
        \cline{2-6}
            & Camera Pose Noise & - & 20 & 65 & 65 & \\
        \cline{2-6}
            & Action Noise & - & 20 & 60 & 60 & \\
        \cline{2-6}
            & Image Noise & - & 20 & 60 & 60 & \\
        \Xhline{4\arrayrulewidth}
    \end{tabular}

    \begin{tablenotes}
        \footnotesize
        \item \textit{Note:} SR stands for Success Rate. Type-weighted SR refers to the success rate weighted by the number of episodes within each type, while study-weighted SR is weighted by the number of episodes within each study.
    \end{tablenotes}
\end{threeparttable}
\end{table*}

 \begin{figure}[!tbp]
  \centering
  \includegraphics[width=\hsize]{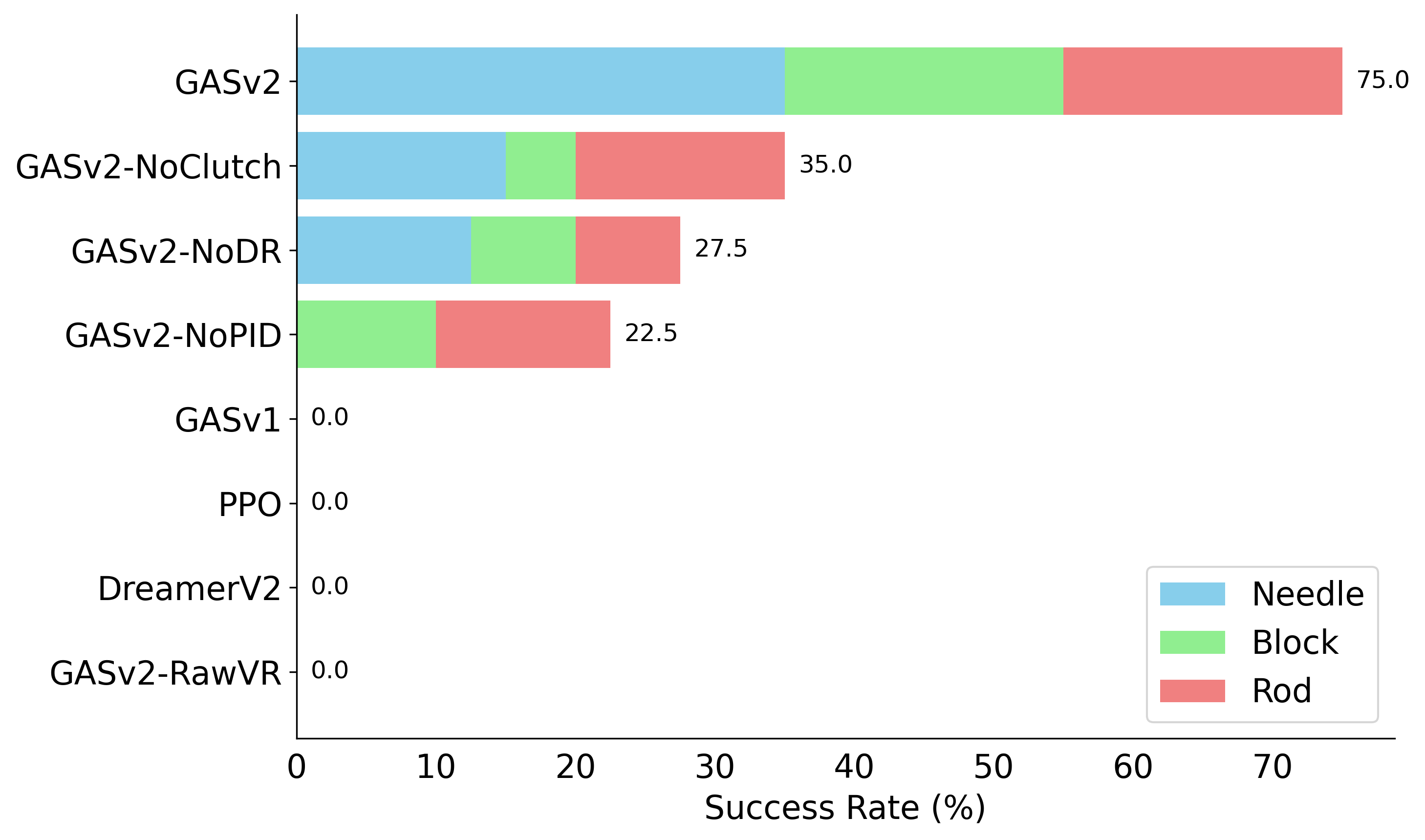}
  \caption{Success rates of performance study in our square phantom across baselines.}
  \vspace{-0.45cm}
  \label{fig:dvrk_baseline_peformance}
\end{figure}

\subsection{Square Phantom}
The square phantom is a silicone platform measuring $10$cm in both width and height.
Target objects are randomly placed on the platform surface for each rollout.
We conduct three categories of experiments on this platform: performance, generality, and robustness.
For the performance study, we use the same target objects as those in the training simulations (see Fig. \ref{fig:surgical_objects}b): a $20$mm needle, a silicone block, and a silicone rod.
The silicone block and rod phantoms are designed to mimic organ tissue in surgical grasping tasks.
To match the occurrence probabilities of these objects with those in the simulation, we evaluate $20$, $10$, and $10$ rollouts for the needle, block, and rod, respectively.
Fig. \ref{fig:dvrk_baseline_peformance} shows the performance comparison across all baselines in the performance study.
Our controller achieves a success rate of $75\%$, significantly outperforming SOTA visuomotor baselines DreamerV2, PPO, and GASv1, all of which achieve zero success rates.
In addition, the ablation baselines show degraded success rates compared to our controller.
Among these, GASv2-NoPID and GASv2-RawVR show success rate reductions of $52.5\%$ and $75.0\%$, respectively, consistent with the trends observed in the simulation results. 
In contrast, GASv2-NoClutch exhibits a significant performance degradation of $40\%$, which was not observed in simulation.
This result highlights the importance of VC in preventing performance loss during sim-to-real transfer.
GASv2-NoDR, which was not evaluated in simulation, also shows a notable performance drop ($47.5\%$) on real robots, highlighting the importance of domain randomization for effective sim-to-real transfer.

As shown in Table \ref{table:dvrk_square_phantom}, our controller achieves over $70\%$ success on all three types of target objects in the performance study.
The success rate of our visuomotor controller shows an improvement of approximately $20\%$ on real robots compared to the simulation results, indicating strong sim-to-real transfer capability.

\begin{table*}[!tbp]
    \centering
    \scriptsize
    \renewcommand{\arraystretch}{1.8}
    \fontsize{8pt}{8pt}\selectfont

    \begin{threeparttable}
        \caption{Real-Robot Experiments on Multi-Level Phantom for GASv2}
        \label{table:dvrk_multilevel_phantom}
        
        \begin{tabular}{|c | c c c | c| c|}
            \Xhline{4\arrayrulewidth}  
            \bf{Study} &\bf{Type} & \bf{Description} & \bf{Episode} & \bf{SR (\%)} & \bf{Study Weighted SR (\%)} \\
            \Xhline{2\arrayrulewidth}
            \multirow{3}{*}{Performance} 
                & Needle &  20mm Needle   & 20 & 60 & \multirow{3}{*}{65} \\
            \cline{2-5}
                & Block & Phantom Block & 10 & 70 & \\
            \cline{2-5}
                & Rod  & Phantom Rod & 10 &  70 & \\
            \Xhline{4\arrayrulewidth}
        \end{tabular}

        \begin{tablenotes}
            \footnotesize
            \item \textit{Note:} SR stands for Success Rate. Study-weighted SR is weighted by the number of episodes within the performance study.
        \end{tablenotes}
    \end{threeparttable}
\end{table*}

To evaluate generality, we evaluate how our controller adapts to the OOD target object, OOD gripper, and OOD camera pose.
For OOD target object, we test five types of target objects (see Fig. \ref{fig:surgical_objects}b): a needle, threads, gauze, a sponge, and debris.
Specifically, we use a $14\ \text{mm}$ needle, which poses a greater challenge due to its smaller size compared to the standard $20$ mm needle.
For the threads, we compare two surgical threads of different lengths: a long thread of $2\ \text{cm}$ and a short thread of $1\ \text{cm}$.
The gauze measures $4\ \text{cm} \times 4\ \text{cm}$, while the sponge is contained within a spherical shape with a diameter of $4\ \text{cm}$.
Lastly, we use a raisin to mimic the debris, enclosed within a sphere of $1\ \text{cm}$ in diameter.
For OOD gripper, we evaluate our controller using an unseen instrument gripper---a forceps gripper (See Fig. \ref{fig:surgical_objects}c), while the target object remains the standard $20\text{mm}$ needle.
For OOD camera pose, we evaluate two unseen camera poses, where the $z$-axis of the camera frame is tilted $0^\circ$ and $40^\circ$ from the ground's vertical.
Notably, the OOD camera poses fall outside the range of the standard $20^\circ$ tilt with camera domain randomization.

In the generality study, our controller achieves an
weighted success rate of $77.5\%$, as Table \ref{table:dvrk_square_phantom} shows.
Among all OOD target objects, grasping threads is the most challenging, achieving a weighted success rate of $65\%$.
For the OOD gripper, performance (success rate: $65\%$) experiences only a slight decline compared to our performance study (success rate: $70\%$), showing a $5\%$ reduction in success rate.
For the OOD camera poses, the $0^\circ$ tilt results in no performance degradation compared to our performance study (success rate: $70\%$), while the $40^\circ$ tilt (success rate: $60\%$) leads to a minor $5\%$ decrease in success rate.
Compared to the performance study, our controller demonstrates no significant degradation, with a $2.5\%$ increase in weighted success rate and only a $5\%$ decrease in the minimum success rate.
These results from the generality study underscore the strong generalization capability of our controller.

\begin{figure}[!tbp]
  \centering
  \includegraphics[width=\hsize]{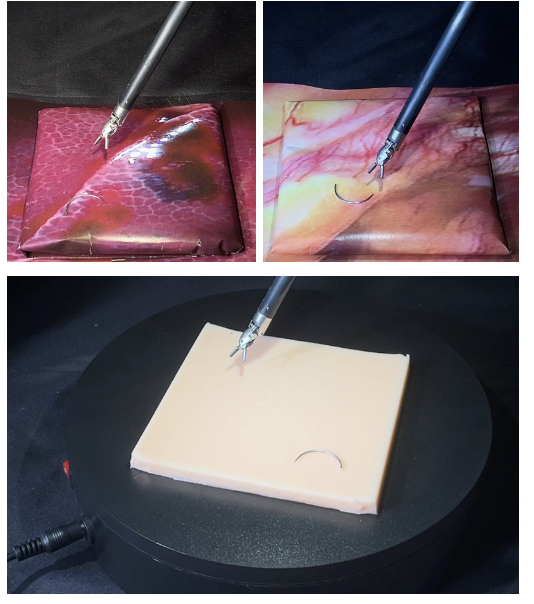}
  \caption{Robustness evaluation of our square phantom platform with two out-of-distribution backgrounds (top) and a motorized rotating platform (bottom).}
  \vspace{-0.45cm}
  \label{fig:backgrounds}
\end{figure}

For the robustness evaluation, we conduct $5$ types of disturbances: background, target noise, camera pose, action noise, and image noise.
To simulate background disturbance, the square phantom is wrapped in two printed papers, each with a different RAS background (see Fig. \ref{fig:backgrounds}).
For target noise disturbance, the square phantom is positioned on a motorized rotating platform, which rotates when actuated, as shown in Fig. \ref{fig:backgrounds}.
As the phantom rotates, the target objects move accordingly, mimicking the motion disturbances observed in real surgical scenarios.
The platform executes a back-and-forth rotation, alternating between clockwise and counterclockwise directions.
This movement induces positional disturbances in the target objects within a range of $2$ cm.
To simulate disturbances in camera pose, noise is added to the calibrated extrinsic camera parameters. These perturbations match those used for camera pose noise in our domain randomization, aiming to emulate calibration errors.
To model action disturbances caused by inaccuracies in robot proprioceptive measurements, random noise is added to the gripper's desired position. This noise is identical to the action noise used in domain randomization.
Similarly, to simulate disturbances in the controller’s input, we apply image noise---identical to that used in domain randomization---directly to the input image.
Importantly, all disturbances in camera pose, action, and input image are unknown to the controller.

Table \ref{table:dvrk_square_phantom} exhibits the performance of our controller in the robustness study.
Our controller achieves a weighted success rate of $61\%$, showing only a minor degradation of $9\%$ compared to the weighted success rate in our performance study.
Among $5$ types of disturbances, our controller achieves a minimum success rate of $55\%$ under target noise disturbances.
The minor performance degradation across these disturbances highlights the strong robustness of our learned controller.

\begin{table*}[!tbp]
\centering
\scriptsize
\renewcommand{\arraystretch}{1.8}
\fontsize{8pt}{8pt}\selectfont

\begin{threeparttable}
    \caption{Real-Robot Experiments on Ex vivo Animal Study for GASv2}
    \label{table:dvrk_exvivo_animal}

    \begin{tabular}{|c | c c | c | c | c|}
        \Xhline{4\arrayrulewidth}  
        \bf{Study}  &\bf{Type} & \bf{Description} & \bf{Episode} & \bf{SR (\%)} & \bf{Study Weighted SR (\%)} \\
        \Xhline{2\arrayrulewidth}

        \multirow{1}{*}{Performance} 
            & Needle & 20mm Needle & 20 & 65 &65 \\
        \Xhline{2\arrayrulewidth}

        \multirow{4}{*}{Generality} 
            & Thread & - & 20 & 70 & \multirow{4}{*}{77.5} \\
        \cline{2-5}
            & Gauze & - & 20 & 95 & \\
        \cline{2-5}
            & Sponge & - & 20 & 90 & \\
        \cline{2-5}
            & Tissue & Porcine Tissue & 20 & 55 & \\
        \Xhline{2\arrayrulewidth}

        \multirow{3}{*}{Robustness} 
            & Background & - & 20 & 65 & \multirow{3}{*}{58.3} \\
        \cline{2-5}
            & Inserted Needle & - & 20 & 55 & \\
        \cline{2-5}
            & Blood Occlusion & - & 20 & 55 & \\
        \Xhline{4\arrayrulewidth}
    \end{tabular}

    \begin{tablenotes}
        \footnotesize
        \item \textit{Note:} SR stands for Success Rate. Study-weighted SR is weighted by the number of episodes within the performance study.
    \end{tablenotes}
\end{threeparttable}
\end{table*}

\subsection{Multi-Level Phantom}

In this study, we investigate whether our controller can grasp objects placed on platforms of varying heights.
We design a platform with varying heights, with maximum height differences of $4\ \text{cm}$.
The top surface of the platform is covered with silicone phantoms.
The width and length of the platform are $10\ \text{cm}$, which are the same as those of the square phantom.
We evaluate the same target objects---the $20\ \text{mm}$ needle, the phantom block, and the rod---used in the performance study on the square platform.
We conduct $20$, $10$, and $10$ episodes for the needle, block, and rod, respectively, matching their occurrence probabilities during training.
The target object is randomly placed on the surface of the platform with height variance to evaluate the success rate.

Table \ref{table:dvrk_multilevel_phantom} presents the performance of our controller on the multi-level platform.
Compared to the square phantom platform, the weighted success rate on the multi-level platform shows a minor degradation of $10\%$.
All target objects experience the same $10\%$ decrease in success rate relative to the square platform.
Among all target objects, grasping the needle remains the most challenging, with a success rate of $60\%$.
This minor performance drop highlights our controller’s ability to robustly grasp objects under height variation.

\subsection{Ex vivo Animal Study}
\label{sec:Ex-vivo Animal Study}
We design experiments to evaluate our controller's effectiveness in ex vivo animal platform.
We use a porcine stomach, measuring approximately $20\text{cm} \times 20\text{cm} \times 5\text{cm}$, as the ex vivo animal platform.
Similar to the studies of the square phantom, we evaluate the performance, generality, and robustness of our controller.

For the performance study, we evaluate our controller by performing 20 rollouts to grasp a standard $20\text{mm}$ needle.
Our controller achieves a success rate of $65\%$, as shown in Table \ref{table:dvrk_exvivo_animal}.
In comparison to the square phantom, the controller used in the ex vivo study demonstrates a modest decrease of $5\%$ in the success rate for the needle picking task.
This highlights the strong performance of our controller on the ex vivo animal platform.

 \begin{figure}[!tbp]
  \centering
  \includegraphics[width=\hsize]{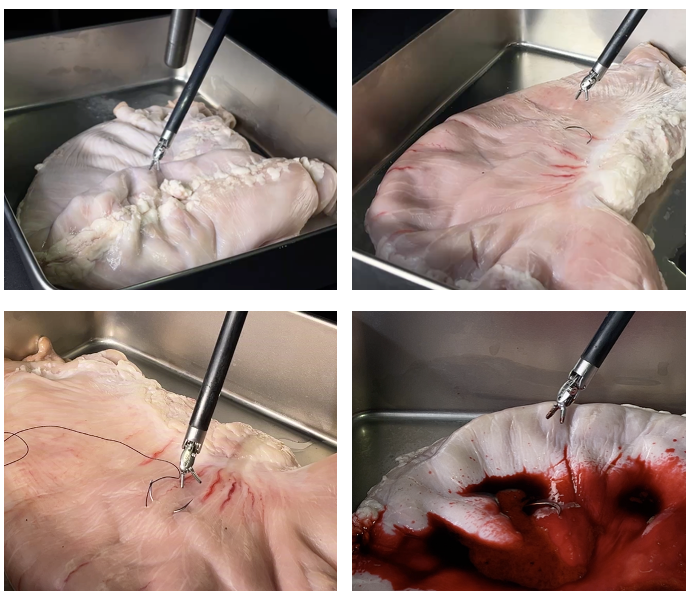}
  \caption{Robustness evaluation in our ex vivo study. The figure shows three disturbance types: background variation (top row), an inserted needle (bottom left), and blood occlusion (bottom right).}
  \vspace{-0.45cm}
  \label{fig:exvivo_robust}
\end{figure}

For the generality evaluation, we evaluate the controller on various target objects, including thread, gauze, sponge, and tissue.
The first three objects (see Fig. \ref{fig:surgical_objects}b) are identical to those used in the evaluation on the square phantom.
For the tissue, we directly use the tissue in the porcine stomach as the target grasping object.
We evaluate $20$ rollouts for each target object.
As Table \ref{table:dvrk_exvivo_animal} shows, our controller achieves a weighted success rate of $77.5\%$.
Among these target objects, the success rate for the tissue is the lowest at $55\%$, indicating the high difficulty of the task.
This may be due to the visual pattern of the tissue as the target grasping object being similar to the background, which poses a significant challenge for the controller’s perception, particularly in image segmentation.
The performance on the thread, gauze, and sponge in the ex vivo study is comparable to the results observed on the square platform.

For the robustness study, we evaluate three types of disturbances (see Fig. \ref{fig:exvivo_robust}): background, inserted needle, and blood occlusion.
For background disturbance, we vary the background by adjusting the platform’s orientation and deforming the morphable organ.
As a result, the background images vary while retaining similar visual patterns.
Notably, we saved mask labels for our visual segmentation and reused them after changing the backgrounds of the platform.
For the disturbance of the inserted needle, the needle is partially inserted into the tissue of the porcine stomach.
This creates visual occlusion and increases the pose variation of the target object.
For the blood occlusion scenario, we partially submerge the needle in blood on the surface of the porcine stomach.
As Table \ref{table:dvrk_exvivo_animal} shows, our controller achieves an average of $58.3\%$ in success rate for the robustness studies.
Among these studies, our controller shows no performance degradation under background disturbances.
This demonstrates strong robustness to disturbances with similar visual patterns, especially considering that object masks were labeled only once.
For the disturbances caused by both the inserted needle and blood occlusion, the controller experiences a modest $10\%$ drop in success rate compared to the needle-picking performance observed in the ex vivo study.
The modest degradation showcases a strong robustness under these two types of disturbance using our controller.

\section{Conclusion and Discussion}

This work introduced \textit{Grasp Anything for Surgery V2} (GASv2), a visuomotor learning framework tailored for surgical grasping tasks. GASv2 combines a world-model-based architecture with a dedicated surgical perception pipeline for processing visual observations, and integrates a hybrid control strategy to ensure safe and reliable execution. The policy is trained entirely in simulation, employing domain randomization to enable robust sim-to-real transfer, and is subsequently deployed on a physical robot in both phantom-based and ex vivo surgical scenarios, using only a single pair of endoscopic cameras.
Comprehensive experiments demonstrate that GASv2 achieves a $65\%$ success rate across both environments, while generalizing effectively to unseen objects and grippers and adapting to diverse disturbances. These results underline the framework’s strong performance, versatility, and robustness, highlighting its potential to advance vision-driven robotic manipulation in surgical contexts.

However, our approach has two primary limitations.
First, users are required to re-annotate object masks in our segmentation module when there is a significant change in the background compared to the original annotations.
Second, the visuomotor controller operates at a relatively low control frequency---approximately 1 Hz---which limits the execution speed of grasping tasks.

In future work, we aim to address these limitations.
To reduce the dependency on manual annotations, we will investigate unsupervised VOS methods for our segmentation module.
Additionally, we aim to explore high-frequency control techniques for visuomotor policies, as suggested in the recent study \cite{intelligence277993634pi0}.

\begin{acks}
We especially thank Ji Woong Kim for sharing valuable insights.
\end{acks}





\bibliographystyle{IEEEtran}
\bibliography{main}

\section*{Appendix A. Experimentation: Accuracy of Depth Estimation}
\begin{figure}[h]
  \centering
  \includegraphics[width=1.0\hsize]{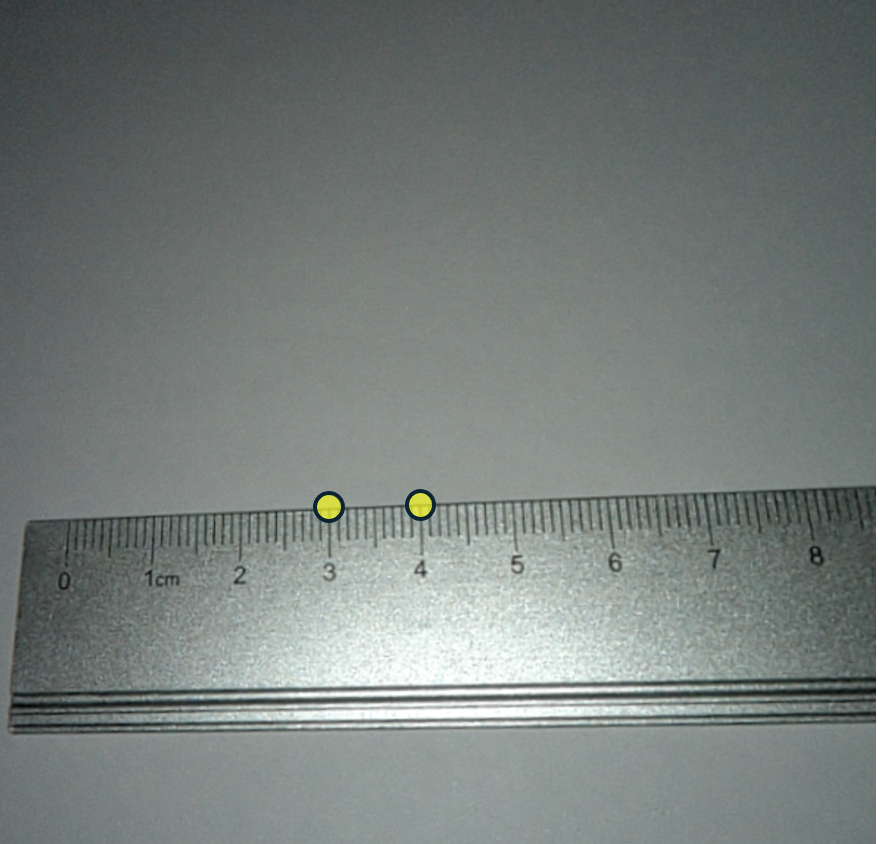}
  \caption{An endoscopic RGB image with annotations for the experiment of depth estimation. The yellow circles are the annotation points for distance $10mm$. Pixel coordinates of the points are saved after annotation.}
  \vspace{-0.45cm}
 \label{fig:depth_accuracy_setup}
\end{figure}

We evaluate our depth estimation by measuring the Cartesian distance between point pairs in the depth image. 
The experimental setup involved positioning the endoscopic stereo camera to capture a top-down view. 
A ruler was placed flat on the ground within the camera's field of view to serve as a reference for depth measurements (see Fig. \ref{fig:depth_accuracy_setup}).
We then manually annotated a pair of feature points on the image, with their ground-truth distance determined using the ruler as a reference.
Fig. \ref{fig:depth_accuracy_setup} shows an example of such a pair, with a known ground-truth distance of $10$ mm.
After the annotation, we record the image coordinates of point pairs. 
For each annotated point pair, we extract a pair of corresponding pixels from the depth image based on their recorded image coordinates.
We project the pixel pair to the Cartesian space based on the camera pinhole model.
The accuracy of depth estimation is evaluated based on the distance error between the projected point distance and the ground-truth distance.
We evaluate point pairs with three ground-truth distances: 10mm, 20mm, and 30mm.
In addition, we assess the effect of varying camera heights, as depth estimation is influenced by the distance between the camera and the feature points, based on the pinhole camera model.
Three camera heights---50mm, 100mm, and 200mm---which are common in the RAS setups, are evaluated.

Table \ref{tab:distance_error_depth_estimation} shows the distance errors over 5 point pairs in our experiments.
For all camera heights and point pairs, the depth error using our depth estimation is all within $1$mm. 
For each camera height, the average error is less than $0.5$mm, with a low standard deviation of no more than $0.55$mm.
As camera height increases, the average errors also rise, consistent with the theoretical predictions of the pinhole camera model.

 \begin{table}[H]
\centering
\scriptsize
\renewcommand{\arraystretch}{1}
\begin{threeparttable}
\caption{Distance Error in Depth Estimation over $5$ Point Pairs}
\begin{tabular}{lcccc}
\toprule
  &  Pair-10mm & Pair-20mm & Pair-30mm & Average \\
\midrule
Height-50mm  &  0.12 (0.42) & 0.14 (0.17) & 0.04 (0.05)  & 0.1 (0.21)   \\
Height-100mm & 0.42 (0.44) & 0.18 (0.38) & 0.13 (0.66) & 0.24 (0.49) \\
Height-200mm & 0.26 (0.98) & 0.44 (0.39) &   0.76 (0.29) &  0.49 (0.55) \\
\bottomrule
\end{tabular}
\label{tab:distance_error_depth_estimation}
\begin{tablenotes}
\footnotesize
\item \textit{Note:} Values are presented as mean (standard deviation). All errors are in millimeters. Ground-truth distances are organized by columns, while the varying heights between the camera and the feature points are listed by rows.

\end{tablenotes}
\end{threeparttable}

\end{table}

\section{Appendix B. Experimentation: Accuracy of Segmentation}
\begin{figure}[h]
  \centering
  \includegraphics[width=1.0\hsize]{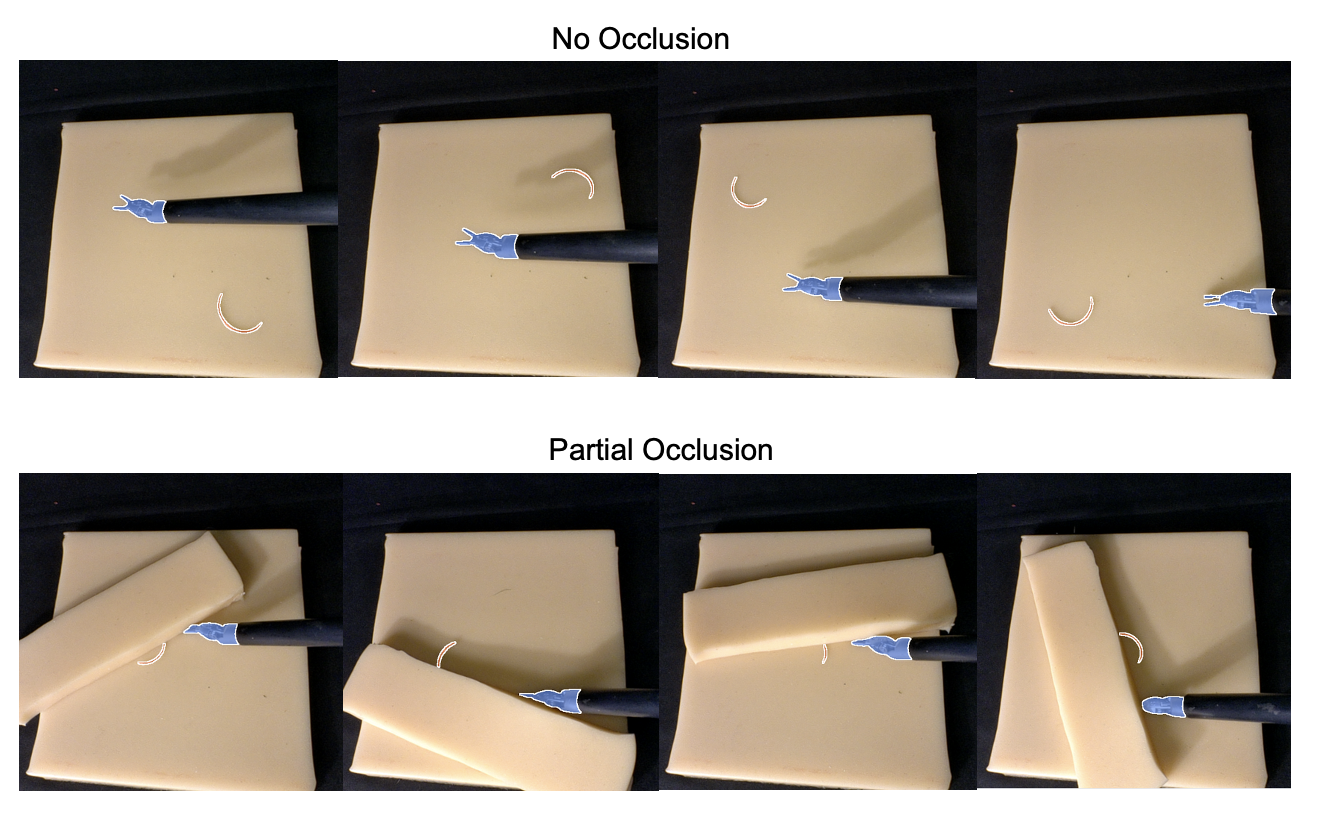}
  \caption{Endoscopic RGB images with mask annotations from our segmentation experiments under different object poses. The top row shows images with no occlusion, while the bottom row presents images with partial occlusion. The orange mask corresponds to a $20$ mm needle, and the blue mask denotes the instrument's gripper.}
  \vspace{-0.45cm}
 \label{fig:mask_accuracy_setup}
\end{figure}

\begin{table}[H]
\centering
\begin{threeparttable}
\caption{Intersection of Union for Mask Tracking}
\begin{tabular}{|c|c|c|}
\hline
\textbf{} & \textbf{Needle (\%)} & \textbf{Gripper (\%)} \\
\hline
\textbf{No Occlusion} & 85.56 (2.24) &  93.15 (0.86) \\
\hline
\textbf{Partial Occlusion} & 83.65 ( 5.41) & 91.82 (1.24) \\
\hline
\end{tabular}
\label{table:mask_tracking_accuracy}
\begin{tablenotes}
\footnotesize
\item \textit{Note:} Values are presented as mean (standard deviation).

\end{tablenotes}
\end{threeparttable}

\end{table}
We conducted experiments to evaluate the accuracy of our segmentation approach using images captured by a stereo endoscopic camera. 
These experiments were designed to replicate the standard robotic-assisted surgery (RAS) setup commonly used for surgical grasping. 
Specifically, the endoscopic camera was mounted on a robotic arm, with its $z$-axis angled $20$ degrees from the vertical plane, perpendicular to the ground.
A surgical robot actuates the instrument's gripper to perform grasping tasks. A $20$ mm needle is placed on a square phantom positioned $15$ cm from the camera. 
The poses of both the needles and gripper are randomized.
Using our segmentation method, we track the masks of both the needle and the instrument's gripper. The experiment is conducted under two occlusion conditions: (i) no occlusion and (ii) partial occlusion. 
For each occlusion condition, we evaluate RGB images with $10$ randomized object poses.
Figure \ref{fig:mask_accuracy_setup} presents RGB images captured by the endoscopic camera under different object poses and occlusion conditions.
For each image, we annotate the object masks using point-prompt annotations from SAM, similar to the process used in our segmentation approach. These annotated masks serve as the ground truth. To evaluate the accuracy of the segmentation, we employ a standard mask metric: the Intersection of Union (IoU).

Table \ref{table:mask_tracking_accuracy} presents the IoU of our mask tracking across different occlusion conditions. Under the no-occlusion condition, both the needle and the gripper demonstrate strong segmentation performance, with IoU values exceeding $85\%$ on average and a small standard deviation of $2\%$. 
In the presence of partial occlusion, the accuracy of mask tracking experiences only a minor degradation, with the mean IoU dropping by less than $2\%$.

\section*{Appendix C. Hyperparameters of GASv2}
\label{sec:appendix_hyperparam}
\begin{table*}[h]
\scriptsize
\renewcommand{\arraystretch}{1.3}
\centering
\begin{tabular}{|>{\raggedright\arraybackslash}m{4cm}|m{4cm}|m{3cm}|m{3cm}|}
\hline
\multirow{2}{*}{\textbf{ Type}} & \multirow{2}{*}{\textbf{ Name}} & \multirow{2}{*}{\textbf{Symbol}} & \multirow{2}{*}{\textbf{Value}} \\
 & & & \\
\hline
\multirow{6}{*}{\textbf{RL Formulation}} 
  & Failed-termination reward &  & -0.1 \\
 & Abnormal-progress reward &  & -0.01\\
 & Normal-progress reward &  & -0.001 \\
 & Translation action scale &  $\alpha^{xyz}$ & 0.3  \\
 & Rotation action scale &  $\alpha^\theta$ & 1.0  \\
 & Time limit & $H_{max}$ & 80 \\
\hline
\multirow{7}{*}{\textbf{Network Architecture and Training}} 
& Replay buffer size &   & $5\times 10^5$ \\
 & Prefill demonstration steps &  & $5\times 10^4$  \\
 & Training steps &  & $1.1\times 10^6$  \\
 & Prefill random steps &  & $5\times 10^4$ \\
 & Batch size &  & 100\\
 & Sequence length  &  & 6 \\
 & RSSM number of units & & 1024 \\
\hline
\multirow{5}{*}{\textbf{Surgical Perception}} &  Number of segememtated task ojects  & $K$ & $3$\\
 & Voxel Resolution & $N_{voxel}$ & $200$ \\
 & DSA zoom range & $N_{zoom}$ & $60$ \\
 & Gripper mask encoding & $e_1$ & $140$ \\
 & Target mask encoding & $e_2$ & $70$ \\
 
\hline
\multirow{6}{*}{\textbf{Hybrid Control Architecture}} 
& VC step & $H_{begin}$ & $6$ \\
 & PID/RL Phase Threshold & $C_{dis}$ & $0.1$ \\
 & Centroid distance $x$-offset & $L_x^{dis}$  & $0$\\
 & Centroid distance $y$-offset & $l_y^{dis}$  & $0$ \\
 & Centroid distance $z$-offset & $l_z^{dis}$  & $0.1$ \\
  & PID proportional gain & $K_p$  & $10$ \\
\hline
\multirow{5}{*}{\textbf{Training Scenes}} 
& Maximum grasp trial &  & $1$\\
 & Needle occurrence &  & $0.5$ \\
 & Block occurrence &  & $0.25$\\
 & Rod occurrence &  & $0.25$ \\
 & Gripper tilt angle &  & $45^{\circ}$ \\
\hline
\end{tabular}
\caption{Hyperparameters for GASv2}
\label{table:hyperparam}
\end{table*}

\end{document}